\documentclass[11pt,letterpaper]{article}
\usepackage{graphicx} % Required for inserting images
\usepackage[utf8]{inputenc}
\usepackage{authblk}

\usepackage[T1]{fontenc}
\usepackage{pbox}
\usepackage{amsmath} %\allowdisplaybreaks[3]
\newcommand{\pushright}[1]{\ifmeasuring@#1\else\omit\hfill$\displaystyle#1$\fi\ignorespaces}
\newcommand{\pushleft}[1]{\ifmeasuring@#1\else\omit$\displaystyle#1$\hfill\fi\ignorespaces}
\usepackage{amsfonts}
\usepackage{amssymb}
\usepackage{bm}
\usepackage{scalerel}
\usepackage[rgb]{xcolor} 
\usepackage{makeidx}
\usepackage{amsmath}
\usepackage{hyperref}
\usepackage{graphicx,import}
\usepackage{gensymb}
\usepackage{epigraph}
\usepackage{caption, subcaption}
\usepackage{calc, csquotes}
\usepackage{algorithm, algpseudocode}
\usepackage{booktabs}% http://ctan.org/pkg/booktabs
\usepackage{siunitx}
\usepackage[version=4]{mhchem}
\usepackage[inline]{enumitem}
\usepackage{wrapfig}
\usepackage{placeins}

\usepackage[margin=0.84in]{geometry}

\newtheorem{remark}{Remark}

\DeclareMathOperator*{\argmax}{argmax}
\DeclareMathOperator*{\argmin}{argmin}
\let\bs\boldsymbol

\title{Multi-Objective Bayesian Optimization for Networked Black-Box Systems: A Path to Greener Profits and Smarter Designs}

\author{Akshay Kudva}
\author{Wei-Ting Tang}
\author{Joel A. Paulson}

\affil{Department of Chemical and Biomolecular Engineering, The Ohio State University, Columbus, 43210, OH, USA}

\begin{document}

\maketitle

\begin{abstract}
Designing modern industrial systems requires balancing several competing objectives, such as profitability, resilience, and sustainability, while accounting for complex interactions between technological, economic, and environmental factors. Multi-objective optimization (MOO) methods are commonly used to navigate these tradeoffs, but selecting the appropriate algorithm (or solver) to tackle these problems is often unclear, particularly when system representations vary from fully equation-based (\textit{white-box}) to entirely data-driven (\textit{black-box}) models. While \textit{grey-box} MOO methods attempt to bridge this gap, they typically impose rigid assumptions on system structure, requiring models to conform to the underlying structural assumptions of the solver rather than the solver adapting to the natural representation of the system of interest. 
In this chapter, we introduce a unifying approach to grey-box MOO by leveraging network (or graph) representations, which provide a general and flexible framework for modeling interconnected systems as a series of function nodes that share various inputs and outputs.
Specifically, we propose MOBONS, a novel Bayesian optimization-inspired algorithm that can efficiently optimize general function networks, including those with \textit{cyclic dependencies}, enabling the modeling of feedback loops, recycle streams, and multi-scale simulations -- features that existing methods fail to capture. Furthermore, MOBONS incorporates constraints, supports parallel evaluations to accelerate convergence, and preserves the sample efficiency of Bayesian optimization while leveraging network structure for improved scalability. We demonstrate the effectiveness of MOBONS through two case studies, including one related to sustainable process design. By enabling efficient MOO under general graph representations, MOBONS has the potential to significantly enhance the design of more profitable, resilient, and sustainable engineering systems.
% For decades, industry thrived on a simple mantra: produce more, faster, and cheaper—often at nature’s expense. Sustainability was an afterthought, regulations were constraints, and optimization meant squeezing out every last ounce of efficiency. But the world has changed. Today, designing resilient and sustainable systems demands more than just brute-force optimization; it requires understanding the intricate web of interactions between technology, economy, and environment. Enter MOBONS—Multi-Objective Bayesian Optimization of Network Systems—a game-changing framework that extends BO beyond traditional black-box and composite models. Unlike existing approaches that assume hierarchical dependencies, MOBONS optimizes cyclic networks, models feedback-driven interactions, and can be easily extended to constrained optimization settings. This unlocks new possibilities for designing sustainable industrial systems that are not only efficient but also ecologically and socially responsible. While MOBONS is a general optimization framework, we focus on its transformative potential in process design for sustainable systems, where balancing competing objectives is critical. By explicitly modeling interdependencies, MOBONS provides a powerful, adaptive tool for shaping the future of responsible engineering.
\end{abstract}

\section{Introduction}

Over the last two centuries, industrial innovation has prioritized efficiency, scale, and profitability, often overlooking the long-term environmental consequences of mass production and resource extraction \cite{boulding1966economics, Bakshi_nature_book}. Early advances in petrochemical refining and chemical manufacturing focused on maximizing throughput and profit margins, with little consideration for pollutant emissions and hazardous waste. As environmental concerns became more pressing, government policies -- such as the Air Pollution Control Act (1955) and the establishment of the Environmental Protection Agency (1970) \cite{EPA_website} -- led to stricter regulations, prompting process systems engineering (PSE) practitioners to integrate environmental and safety constraints into their design objectives. By the 1970s, nonlinear programming (NLP) methods were being used to optimize capital cost objectives under uncertainty \cite{Takamatsu1970, Nishida1974}, followed by advances in feasibility and flexibility analysis \cite{Grossmann1978, Halemane1983, Swaney1985, Grossmann1987} and later integrated design and control (IDC) frameworks \cite{Narraway1991, Luyben1994, Burnak2019}. These developments laid the foundation for more holistic optimization approaches that simultaneously consider economic, operational, and regulatory factors.

In parallel, the increasing availability of computational resources enabled the rise of high-fidelity simulation tools, commonly referred to as \textit{digital twins} \cite{Niederer2021, Sharma2022}, which model complex physical and chemical systems with greater accuracy. However, this increase in model fidelity has come at the cost of \textit{computational expense}, making purely equation-oriented (i.e., ``white-box'') approaches less tractable for large-scale, multi-physics systems. At the same time, the rise of data-driven machine learning and Bayesian optimization (BO) \cite{Kushner1964, Mokus1975} has provided an alternative by enabling the efficient optimization of expensive, black-box processes. BO has proven particularly valuable for optimizing systems where the internal structure is either proprietary, analytically intractable, or hidden behind computationally expensive simulations. This has led to successful applications in process design, reaction optimization, and flexibility analysis \cite{Hickman2022, Kudva2022,banerjee2010computationally, rogers2015feasibility, Geremia2023, Kudva2024, Paulson2024BO4Sustaibalilty}.

\begin{figure}[tb]
  \centering
  \includegraphics[width=0.9\textwidth]{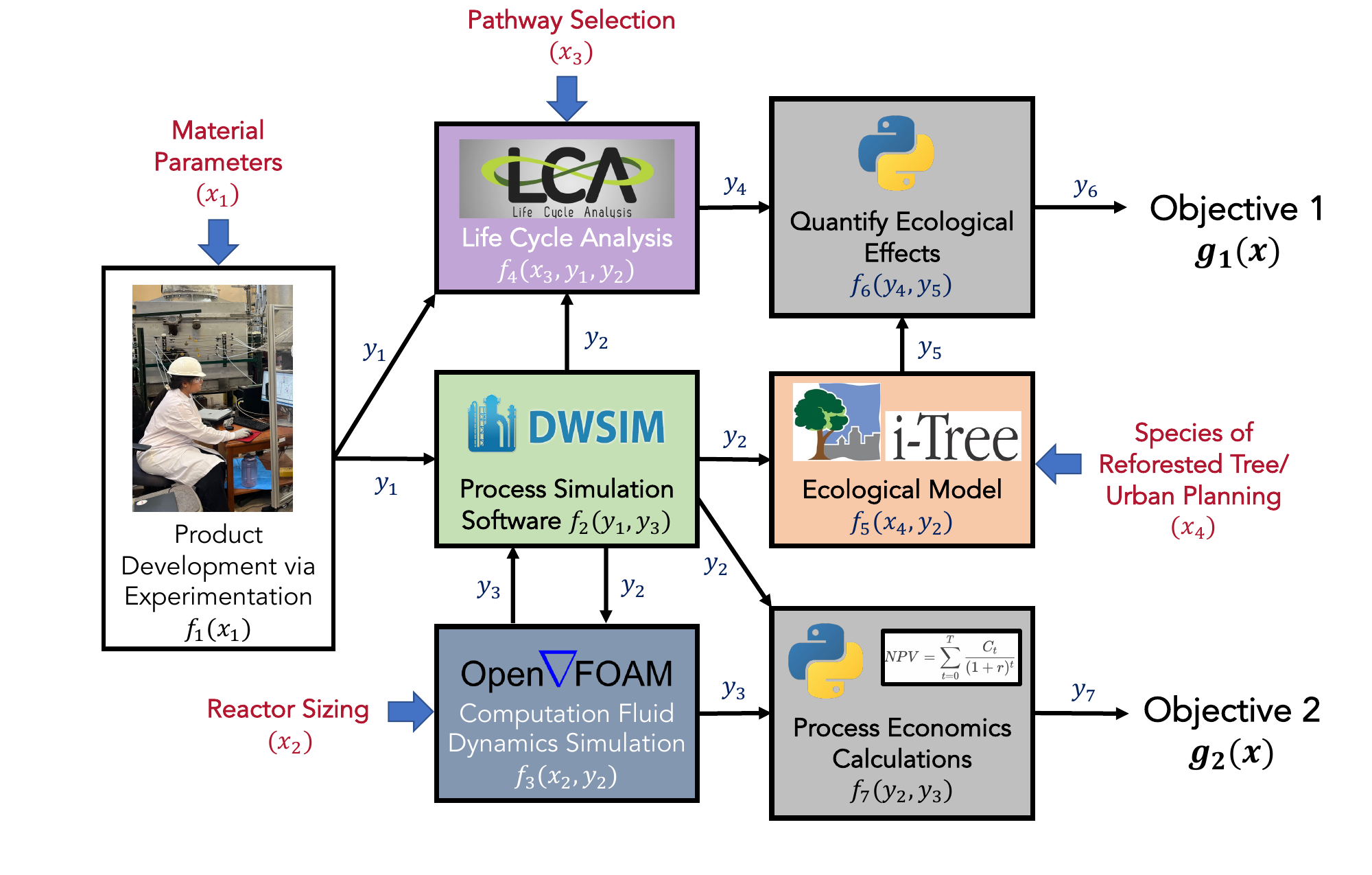}
  \caption{Illustration of a complex network system integrating process simulation \cite{DWSIM}, CFD \cite{openfoam}, life cycle analysis (LCA), ecological modeling \cite{iTree}, and economic evaluation. The design variables (highlighted in red) impact reactor sizing, material properties, and urban planning, while the models interact dynamically to assess environmental and economic trade-offs. The figure serves as a motivating example, highlighting the challenges inherent in optimizing such systems. We aim to develop algorithms capable of tackling a broad class of these problems, which often involve high-fidelity partial differential equation solvers, embedded optimization modules, and dynamic feedback loops.}
  \label{fig:Ecosystem_function_network}
\end{figure}

Despite BO's successes, treating an entire process as a black-box often discards valuable structural information that could significantly improve optimization efficiency, especially as the number of decision variables increases. Traditional equation-oriented methods exploit model structure but require access to explicit equations -- a luxury not always available due to proprietary software or the complexity of modern simulators. To bridge this gap, ``grey-box'' (Bayesian) optimization methods have been developed within the PSE community \cite{Boukouvala2016, Beykal2018, paulson2022cobalt, Kudva2022draco, Schweidtmann2021, Lu2023, lu2025bo4io}, which integrate known submodels with data-driven surrogates. While these methods improve efficiency over purely black-box BO, they usually assume a sequential or hierarchical mapping from inputs to outputs, which may not be representative of many real-world systems. 
Many engineering systems, particularly in PSE, exhibit networked dependencies, where subsystems interact in a complex, often bidirectional manner. Figure~\ref{fig:Ecosystem_function_network} illustrates such a scenario, where a process system is coupled with environmental and economic subsystems. Here, high-fidelity tools such as computational fluid dynamics (CFD) models \cite{openfoam}, life cycle analysis (LCA) \cite{Guinee2002}, ecological models \cite{iTree}, and process simulators \cite{DWSIM} interact dynamically, often in a non-hierarchical fashion. A multi-scale simulation -- such as one linking a flowsheet model to a detailed heat exchanger CFD simulation -- may require back-and-forth data exchange, creating cycles that challenge many existing optimization methods. 
A naive black-box BO approach would ignore these interconnections, potentially requiring an unmanageable number of function evaluations. Conversely, a strictly equation-oriented approach might be infeasible if key subsystems are black-boxes due to proprietary software restrictions or excessive modeling complexity. This necessitates a more flexible yet structure-aware optimization framework.

Recent work in so-called ``composite'' BO has sought to extend BO to systems where one function’s output feeds into another \cite{astudillo2019bayesian, paulson2022cobalt, Lu2023}. However, these methods assume that the function can be expressed as a composite model, $g(\bs{h}(\bs{x}))$, where $g$ is known, $\bs{h}$ is a black-box function, and $\bs{x}$ are the design variables. A more general representation of the function would be as a network (or graph), which has been explored in recent BO frameworks such as BOFN \cite{BOFN, BOFN_PE} and BOIS \cite{gonzalez2024bois, gonzalez2025implementation}. Graph-based models are known to be powerful representations are seen by their wide applicability in many different fields such as social sciences, biology, chemistry, and engineering \cite{khoshraftar2024survey}. However, these current graph-based BO methods assume that the function network is a directed acyclic graph (DAG) to simplify the uncertainty propagation task. This assumption limits their applicability to process systems with feedback loops, such as recycle streams and interconnected multi-scale simulations.

In this chapter, we extend MOBO to handle complex networked systems, where each node in the network represents a potentially expensive black-box function (e.g., a simulator or data-driven model). We introduce Multi-Objective Bayesian Optimization for Network Systems (MOBONS), which builds upon existing methods while addressing key limitations:
\begin{itemize}
    \item MOBONS handles multi-objective optimization problems, seeking Pareto-optimal solutions that balance competing objectives such as economic performance and sustainability.
    \item It explicitly models networked dependencies, including feedback loops and recycle streams, enabling applications in complex industrial systems.
    \item It naturally accommodates constraints (e.g., safety or regulatory requirements) by treating them as additional outputs of the function network. 
    \item It is readily applicable to batch (parallel) evaluations such that it can accelerate the optimization process by designing multiple candidate evaluation points at each iteration. 
\end{itemize}
To ensure computational efficiency of the algorithm, MOBONS extends the Thompson sampling acquisition function \cite{kandasamy2018parallelised}, which is significantly cheaper than the improvement-based approach in BOFN \cite{BOFN} and the confidence bound approach in BOIS \cite{gonzalez2024bois, gonzalez2025implementation}. Furthermore, we leverage multi-objective genetic algorithms to efficiently solve the subproblems within MOBONS, combining the sample efficiency of BO with the exploration power of genetic search. Note that we have also recently pursued related work in the context of function networks under adversarial uncertainty (i.e., robust optimization) \cite{kudva2025bonsai}, though we do not directly consider that case in this work. 

The remainder of this chapter is organized as follows. Section~\ref{sec:problem-formulation} formalizes the problem statement. Section~\ref{sec:MOBONS} presents the MOBONS framework, including surrogate modeling and acquisition strategies. Section~\ref{sec:case-study} demonstrates the method on case studies. Finally, Section~\ref{sec:conclusions} provides concluding remarks.

\section{Problem Formulation} \label{sec:problem-formulation}

In this chapter, we consider the problem of multi-objective optimization in the presence of expensive function evaluations. Formally, we seek to solve the following optimization problem
\begin{align} \label{eq:moo}
    \min_{\bs{x} \in \mathcal{X}} ~ \bs{G}(\bs{x}) = [ g_1(\bs{x}), g_2(\bs{x}), \ldots, g_M(\bs{x}) ],
\end{align}
where $\mathcal{X} \subset \mathbb{R}^D$ is a compact design space, $\bs{x}$ is the decision vector representing all the degrees of freedom in the system, and $\bs{G} : \mathcal{X} \to \mathbb{R}^M$ is a vector-valued function consisting of $M$ scalar objectives $g_i : \mathcal{X} \to \mathbb{R}$, each of which quantifies a cost or performance metric to be minimized. While the objective functions in \eqref{eq:moo} can represent arbitrary criteria, we are motivated by cases where at least one objective relates to economic considerations (e.g., net present value, return on investment), while others capture environmental sustainability metrics (e.g., pollutant emissions, energy consumption, waste generation). 

Multi-objective optimization problems of this form typically do not admit a single globally optimal solution, as improving one objective may degrade another. Instead, the goal is to construct the set of \textit{Pareto-optimal solutions}, forming the Pareto frontier, where no objective can be improved without compromising at least one other. The objective of this work is to develop an algorithm that enables efficient construction of the Pareto frontier by leveraging the underlying structure of $\bs{G}$, which is often represented as a network of interconnected functions.

The choice of optimization algorithm for solving \eqref{eq:moo} depends on the properties of $\bs{G}$. The $\epsilon$-constraint method \cite{mavrotas2009effective} is effective when each single-objective subproblem can be efficiently minimized while imposing constraints on the remaining objectives; sweeping over the constraint parameters then traces out the Pareto frontier. However, this approach requires access to a fully specified model that can be solved using an optimization modeling framework such as Pyomo \cite{hart2011pyomo}. When the objective functions are treated as black-box evaluations, evolutionary algorithms like NSGA-II \cite{Deb2002} provide an alternative that does not require explicit function derivatives or structural knowledge. While broadly applicable, these methods require a large number of function evaluations and are thus impractical for expensive-to-evaluate functions, such as those arising from high-fidelity process simulations.
A more sample-efficient alternative is multi-objective Bayesian optimization (MOBO), which sequentially selects evaluation points based on a surrogate model trained on previously observed data. Popular MOBO strategies include expected hypervolume improvement (EHVI) \cite{Daulton2020} and Thompson sampling-based methods, such as TSEMO \cite{Bradford2018} and POTS \cite{Ranganathan2023}. While these approaches significantly reduce the number of function evaluations, they still treat $\bs{G}$ as a fully black-box function, which can limit optimization performance in engineering applications. We argue that this representation is often overly flexible and overlooks valuable structural information that is readily available in many real-world problems.

\begin{figure}[tb]
  \centering
  \includegraphics[width=0.9\textwidth]{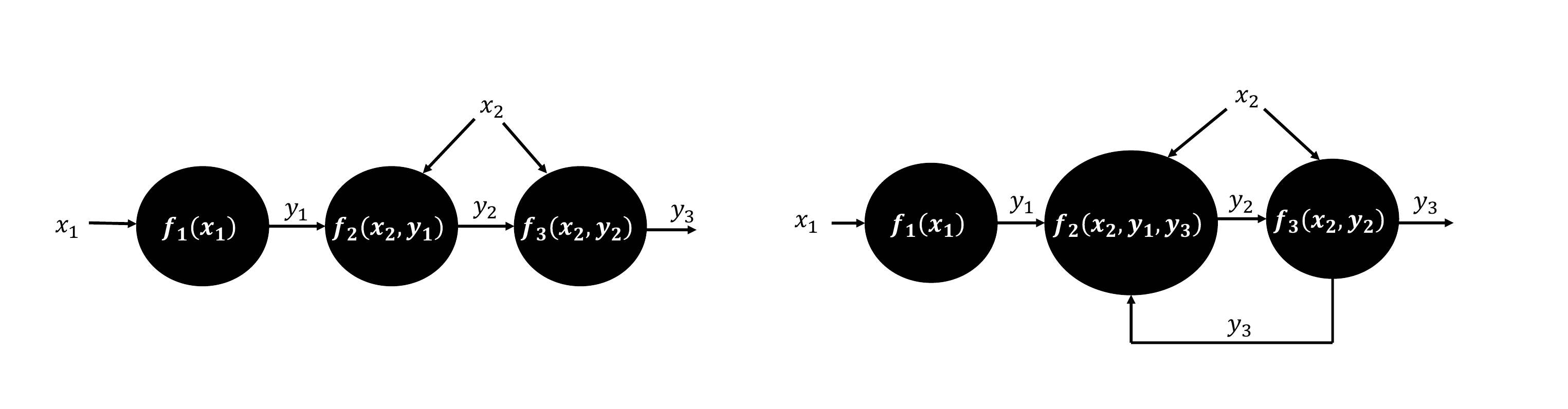}
  \caption{Illustrative examples of acyclic (left) and cyclic (right) function networks. The acyclic graph must be capable of being topologically ordered so that future nodes only depend on previous nodes, which simplifies the evaluation. The cyclic graph, on the other hand, may have dependencies that come in the form of equality constraints, which significantly complicates determination of statistical properties of the network when the node functions are modeled as Gaussian processes.}
  \label{fig:cyclic_vs_acyclic}
\end{figure}

Building on recent work on function networks \cite{BOFN}, we consider a structured representation of $\bs{G}$ in which the objectives are evaluated through a directed network of interdependent functions $f_1, \ldots, f_K$. Each function $f_k$ in the network receives inputs from a subset of external design variables $\bs{x}$ as well as the outputs of other functions in the network. The evaluation of these functions is assumed to be expensive, meaning that computational or experimental resources are required for function evaluations. 
To formalize this structure, let $V = \{ 1, 2, \ldots, K \}$ represent the set of nodes in the network, where each node corresponds to a function $f_k$. The directed edges of the network, denoted by $E = \{ (j, k) : f_k \text{~receives input from~} f_j \}$, specify the dependencies between functions. Each function $f_k$ can be expressed as an implicit equation
\begin{align} \label{eq:network-functions}
    y_k = f_k( \bs{x}_{I(k)}, \{ y_{j} \}_{j \in J(k)} ), \quad k = 1,\ldots, K,
\end{align}
where $\bs{x}_{I(k)}$ represents the subset of decision variables that directly influence $f_k$, with $I(k) \subseteq \{1,\dots,D\}$ and $J(k) \subseteq \{1,\dots,K\}$ is the set of other functions whose outputs serve as inputs to $f_k$. Let $\bs{Y} = (y_1, y_2, \dots, y_K)$ collects all node outputs in the network.
For compactness, we define the vector-valued function $\bs{F}(\bs{x}, \bs{Y})$ as the concatenation of the right-hand sides of \eqref{eq:network-functions}, allowing us to rewrite \eqref{eq:network-functions} as
\begin{align} \label{eq:compact-network}
    \bs{Y} = \bs{F}(\bs{x}, \bs{Y}).
\end{align}
This formulation is more general than prior work that assumes an acyclic network structure. Here, we explicitly allow cycles, enabling the representation of feedback loops and recycle streams, which are common in process systems applications. Figure \ref{fig:cyclic_vs_acyclic} provides illustrative examples of cyclic and acyclic function networks to highlight their differences. 

The objective function $\bs{G}(\bs{x})$ is then assumed to be a known transformation of the network outputs. Without loss of generality, we assume a linear transformation of the form
\begin{align} \label{eq:objective-matrix}
    \bs{G}(\bs{x}) = \bs{C} \bs{Y}^\star(\bs{x}),
\end{align}
where $\bs{C} \in \mathbb{R}^{M \times K}$ is a projection matrix and $\bs{Y}^\star(\bs{x})$ represents the solution to the implicit system \eqref{eq:compact-network} for a given $\bs{x}$. We do not strictly assume $\bs{Y}^\star(\bs{x})$ is unique for every $\bs{x}$, though this is expected to improve numerical stability in practice. An interesting direction for future work is to study the impact of properties of the solution map on performance. A key distinction between our approach and conventional MOBO is how the surrogate model is defined. Instead of placing a Gaussian process prior directly on the objective functions $g_1, \dots, g_M$, we propose placing one on the network functions $f_1, \dots, f_K$. While this may appear to be a minor change, it allows us to explicitly exploit the dependency structure in \eqref{eq:network-functions}, which is ignored in standard MOBO methods. By leveraging this network representation, we aim to improve the efficiency of multi-objective optimization in expensive evaluation settings.

In the next section, we describe our proposed approach for incorporating this structured information into a MOBO-like framework.

\section{Multi-Objective Bayesian Optimization over Network Systems} \label{sec:MOBONS}

In this section, we introduce our approach for incorporating network structure into the multi-objective optimization of expensive functions. We first describe the statistical model used to represent the black-box node functions $f_1, \ldots, f_K$, leading to a fundamentally different modeling approach than standard MOBO. We then present the proposed MOBONS algorithm, detailing its key components and workflow. Finally, we discuss extensions that enable parallel evaluations and the handling of inequality constraints.

\subsection{Statistical surrogate model}

We assume that each function $f_k$ is a realization from an independent Gaussian process (GP) prior. To simplify notation, we define $\bs{z}_k = (\bs{x}_{I(k)}, \bs{Y}_{J(k)})$ as the vector of inputs to node $k$, where $\bs{x} \in \mathbb{R}^D$ represents the external decision variables and $\bs{Y} \in \mathbb{R}^K$ collects the outputs of all nodes. Each function $f_k$ is then modeled as
\begin{align}
    f_k(\bs{z}_k) \sim \mathcal{GP}(\mu_{0,k}(\bs{z}_k), \Sigma_{0,k}(\bs{z}_k, \bs{z}_k')),
\end{align}
where $\mu_{0,k}(\bs{z}_k) : \mathbb{R}^{n_{z,k}} \to \mathbb{R}$ is the prior mean function and $\Sigma_{0,k}(\bs{z}_k, \bs{z}_k') : \mathbb{R}^{n_{z,k}} \times \mathbb{R}^{n_{z,k}} \to \mathbb{R}_+$ is the prior covariance (or kernel) function (note $n_{z,k}$ denotes the dimension of $\bs{z}_k$ and $\mathbb{R}_+$ denotes the set of positive real numbers). The mean function encodes prior knowledge about the expected behavior of $f_k$, while the kernel function defines correlations between function values at different inputs, capturing smoothness and variability. A detailed review of kernel choices can be found in \cite[Chapter 4]{williams2006gaussian}; in this work, we assume an appropriate kernel is selected, potentially learned from data. 

We assume that, when the objective functions $\bs{G}$ are evaluated at a design point $\bs{x}$, the intermediate node outputs $\bs{Y}$ are also observed. This assumption holds in many practical settings, such as high-fidelity simulations where intermediate process states can be accessed. Under this assumption, we can infer the value of $f_k(\bs{z}_k^\star)$ at an arbitrary test input $\bs{z}_k^\star$ using a GP posterior conditioned on $n$ previously observed function evaluations. Let $\mathcal{D}_{n,k} = \{ (\bs{z}_{i,k}, y_{i,k}) \}_{i=1}^{n}$ denote the dataset of past evaluations for node $k$. The posterior mean and covariance functions for $f_k$ are given by
\begin{subequations} \label{eq:posterior-gp}
\begin{align}
    \mu_{n,k}(\bs{z}^\star_k) &= \mu_{0,k}(\bs{z}^\star_k) + \Sigma_{0,k}( \bs{z}^\star_k, \bs{z}_{1:n, k} ) \Sigma_{0,k}( \bs{z}_{1:n, k}, \bs{z}_{1:n, k} )^{-1} ( y_{1:n, k} - \mu_{0,k}( \bs{z}_{1:n, k} ) ), \\
    \Sigma_{n,k}(\bs{z}^\star_k, \bs{z}^{\star '}_k) &= \Sigma_{0,k}(\bs{z}^\star_k, \bs{z}^{\star '}_k) - \Sigma_{0,k}( \bs{z}^\star_k, \bs{z}_{1:n, k} ) \Sigma_{0,k}( \bs{z}_{1:n, k}, \bs{z}_{1:n, k} )^{-1} \Sigma_{0,k}(\bs{z}_{1:n, k}, \bs{z}^{\star '}_k).
\end{align}
\end{subequations}
The posterior distributions on $f_1,\ldots,f_K$ induce a corresponding posterior on $\bs{G}$, but this distribution is generally non-Gaussian and difficult to characterize. When the function network is \textit{acyclic}, samples from $\bs{G}$ can be obtained via a forward propagation procedure, as described in \cite[Algorithm 1]{BOFN}. However, when the graph contains \textit{cycles}, the outputs $\bs{Y}$ are defined implicitly by the stochastic fixed-point equation \eqref{eq:compact-network}. In such cases, drawing samples requires numerical techniques such as stochastic fixed-point iteration, Markov Chain Monte Carlo (MCMC), or variational inference, which complicates integration into a sequential optimization framework. We address this challenge by leveraging an efficient heuristic from the multi-armed bandit literature: Thompson sampling (TS) \cite{kandasamy2018parallelised}.

\begin{remark}
    While we assume that all functions in the network are expensive-to-evaluate black-box functions modeled as Gaussian processes (GPs), in practice, some node functions may be inexpensive and explicitly known (i.e., white-box functions). This scenario can be easily accommodated by directly specifying the known function values in the prior. Specifically, for any white-box function $f_k$, we can set the prior mean function as $\mu_{0,k}(\bs{z}_k) = f_k(\bs{z}_k)$ and the covariance function as $\Sigma_k(\bs{z}_k, \bs{z}_k') = 0$, ensuring that the surrogate model exactly recovers the known function values while maintaining uncertainty quantification for the remaining black-box functions.
\end{remark}

\subsection{Multi-objective Thompson sampling for function networks}

The TS method selects an action (in this case, the next design vector $\bs{x}_{t+1}$) based on its probability of being optimal. At iteration $t$, the next query point is sampled from the posterior distribution
\begin{align} \label{eq:TS-single-objective}
    p_{\bs{x}^\star}(\bs{x} | \mathcal{D}_t) = \int \delta\left( \bs{x} - \argmin_{\bs{x} \in \mathcal{X}} g(\bs{x}) \right) p( g | \mathcal{D}_t ) dg,
\end{align}
where $g$ is the single objective function, $\mathcal{D}_t$ is the history of query-observation pairs up to step $t$, and $\delta(\cdot)$ is the Dirac delta function, meaning all probability mass is placed on the minimizer of a sampled function realization. Thus, $\bs{x}_{t+1} = \argmin_{ \bs{x} \in \mathcal{X} } \hat{g}_t(\bs{x})$ where $\hat{g}_t(\bs{x}) \sim p( g | \mathcal{D}_t )$ is a function realization from the GP posterior.
This idea has been extended to MOBO by sampling from the probability that a design point is Pareto optimal \cite{Ranganathan2023}. However, existing methods do not account for the network structure in \eqref{eq:network-functions}. 

Our approach, which we refer to as MOBONS, does account for this structure, as shown in Algorithm \ref{alg:mobons}. The algorithm proceeds as follows. We start by collecting initial data from the network that is used to train the GP models (Lines 1 and 2). We then enter a loop (Line 3) to sequentially generate the remaining set of samples until our total budget $N$ is exhausted. At any given iteration $t$, we solve a multi-objective optimization problem inspired by the TS approach (Lines 4 to 7). In particular, we can interpret \eqref{eq:acquisition-func} as generating points according to the probability that they are Pareto optimal, i.e., 
\begin{align}
    & p_{X^\star, Y^\star}( \bs{x}, \bs{Y} | \{ \mathcal{D}_{t,k} \}_{k=1}^K) \\\notag 
    & = \int \mathbb{I}\left( (\bs{x}, \bs{Y}) \in \argmin_{\bs{x}\in\mathcal{X},\bs{Y}} \{ \bs{C}\bs{Y} : \bs{Y} = \bs{F}(\bs{x}, \bs{Y}) \} \right)p( f_1 | \mathcal{D}_{t,1} ) \cdots p( f_K | \mathcal{D}_{t,K} ) df_1 \cdots df_K,
\end{align}
where $\mathbb{I}(\bs{q} \in Q)$ is the indicator function for set $Q$. 
Note that $(X^\star_{t+1}, Y^\star_{t+1})$ from \eqref{eq:acquisition-func} are the Pareto sets for the TS version of the network function and thus themselves could contain an infinite set of points in general. As such, we need some selection criteria to pick a single point from these sets to evaluate (Line 8; we discuss the generalization to multi-point selection in Section \ref{subsec:parallel-constraints}). We leave the selection function $\Phi(\cdot)$ open since, in principle, any choice from $(X^\star_{t+1}, Y^\star_{t+1})$ is Pareto optimal for the drawn function samples. A simple strategy is to randomly select a point from this set; however, to promote stronger diversity among the candidate points, it is common to select points that are the furthest away from our past samples, e.g., \cite{sun2019synthesizing}. Due to the network structure, we can select points globally (those that maximize distance to the global set of design variables $\bs{x}$) or locally (those that maximize distance to the input to each local node $k$). For simplicity, we focus on the global distance in this chapter to serve as a proof of concept -- future work should more systematically explore the influence of different choices of selection strategies. 
To define the global selection approach $\Phi(\cdot)$, let us first define the selected design point from $X_{t+1}^\star$ to be the one that has the largest \textit{maximin} distance from all previous selected candidates $X_t$:
\begin{align} \label{eq:global-selection}
    \bs{x}_{t+1} = \argmax_{\bs{x} \in X_{t+1}^\star} \min_{\bs{x}_i \in X_t} \|  \bs{x} - \bs{x}_i \|.
\end{align}
For this choice of $\bs{x}_{t+1}$, we can define the corresponding output values $\bs{Y}_{t+1}$ as those satisfying the TS network equations for this choice of design variables, i.e., $\bs{Y}_{t+1} = \widehat{\bs{F}}_{t}(\bs{x}_{t+1}, \bs{Y}_{t+1})$. The specific input values for every node $k = 1,\ldots,K$ are then defined by $\bs{z}_{t+1,k} = ( \bs{x}_{t+1, I(k)}, \bs{Y}_{t+1, J(k)} )$ (Line 8). Lastly, the final steps (Lines 9--13) require us to query the nodes of the network function and update the GP posterior with this newly collected data. 

\begin{algorithm}[tb!]
\caption{MOBONS: Multi-Objective Bayesian Optimization over Network Systems}
\textbf{Required:} Design space $\mathcal{X} \subset \mathbb{R}^D$; 

~~~~~~~~~~~~~~~ Network output space $\mathcal{Y} \subset \mathbb{R}^K$; 

~~~~~~~~~~~~~~~ GP priors $f_k \sim \mathcal{GP}( \mu_{0,k}, \Sigma_{0,k} )$ for all $k = 1,\ldots,K$; 

~~~~~~~~~~~~~~~ Graph representation $(V, E)$ of \eqref{eq:network-functions};

~~~~~~~~~~~~~~~ Matrix $\bs{C} \in \mathbb{R}^{M \times K}$ that projects network outputs to objectives;

~~~~~~~~~~~~~~~ Function $\Phi(\cdot)$ to select candidate points from Pareto set;

~~~~~~~~~~~~~~~ Initialization budget $n$;

~~~~~~~~~~~~~~~ Total evaluation budget $N$.

\begin{algorithmic}[1]
\State \textbf{Initialize data:} For every $k = 1,\ldots,K$, generate initial set of $n$ sampling points $\bs{z}_{1:n, k}$ using any space filling design (such as Latin hypercube sampling) inside of the compact set $\mathcal{Z}_k = \mathcal{X}_{I(k)} \times \mathcal{Y}_{J(k)}$. Evaluate the function $f_k$ at these points via \eqref{eq:network-functions} to construct initial dataset $\mathcal{D}_{n,k} = \{ (\bs{z}_{i,k}, y_{i,k}) \}_{i=1}^n$. 
\State \textbf{Train GPs:} Train GP surrogate for $f_k$ using $\mathcal{D}_{n,k}$ for all $k = 1,\ldots, K$. Let $\mathcal{GP}_{n,k} \leftarrow \mathcal{GP}( \mu_{n,k}, \Sigma_{n,k} )$ denote the GP posterior for $f_k$ that can be evaluated via \eqref{eq:posterior-gp}. 
% (includes optional hyperparameter tuning step to train parameters in the mean and/or covariance functions)
\For{$t=n, n+1, \ldots, N-1$}
\For{$k=1,\ldots,K$}
\State Generate a Thompson sample $\widehat{f}_{t,k} \sim \mathcal{GP}_{t,k}$. 
\EndFor
% \State For all $k = 1,\ldots, K$, 
\State Solve the following multi-objective optimization problem:
\begin{align} \label{eq:acquisition-func}
    (X^\star_{t+1}, Y^\star_{t+1}) = \argmin_{(\bs{x}, \bs{Y}) \in \mathcal{X} \times \mathbb{R}^K} \bs{C}\bs{Y} ~~ \text{subject to} ~~ \bs{Y} = \widehat{\bs{F}}_t(\bs{x}, \bs{Y}),
\end{align}
where $\widehat{\bs{F}}_t$ denotes the compact representation of the network (see \eqref{eq:compact-network}) constructed with the Thompson sample versions of $\widehat{f}_{t,k}$ instead of the true (unknown) $f_k$ and $(X^\star_t, Y^\star_t)$ denotes the set of Pareto optimal design and network output values. 
\State Apply selection rule $\Phi(\cdot)$ to Pareto set $\{ \bs{z}_{t+1, k} \}_{k=1}^K = \Phi( X_{t+1}^\star, Y_{t+1}^\star )$ to get promising candidates. 
\For{$k=1,\ldots,K$}
% \State Select the sample with largest \textit{maximin} distance to the previously observed input values:
% \begin{align}
%     \bs{z}_{t+1, k} = \argmax_{\bs{z}^\star \in Z_{t, k}^\star} \min_{(\bs{z}_{i,k}, y_{i,k}) \in \mathcal{D}_{t,k}} \gamma( \bs{z}^\star, \bs{z}_{i,k} ),
% \end{align}
% where $\gamma(\cdot)$ is some distance function (e.g., Euclidean distance) and $Z_{t, k}^\star = X^\star_{t, I(k)} \times Y^\star_{t, J(k)}$ is the subset of Pareto optimal input values for function node $k$. 
\State Evaluate $f_k$ at $\bs{z}_{t+1, k}$ and update dataset $\mathcal{D}_{t+1,k} \leftarrow \mathcal{D}_{t,k} \cup \{ (\bs{z}_{t+1, k}, y_{t+1,k}) \}$. 
\State Update GP posterior using new data $\mathcal{GP}_{t+1,k} \leftarrow \mathcal{GP}( \mu_{t+1,k}, \Sigma_{t+1,k} )$. See \eqref{eq:posterior-gp}.
\EndFor
\EndFor
\end{algorithmic}
\label{alg:mobons}
\end{algorithm}

\subsection{Practical considerations in MOBONS}

The conceptual MOBONS approach (Algorithm \ref{alg:mobons}) involves several internal steps that require further elaboration for practical implementation.

\paragraph{GP kernel selection and tuning.} The choice of covariance functions (kernels) in the GP priors is critical in MOBONS, as it governs the properties of the surrogate models used to fit the unknown functions. For example, the radial basis function (RBF) kernel, given by
\begin{align}
    \Sigma_{\text{RBF}}( \bs{z}, \bs{z}' ) = \zeta^2 \exp\left( -\frac{\| \bs{z} - \bs{z}' \|^2}{2\ell^2} \right),
\end{align} 
encodes the prior assumption that function values closer in the input space should be more correlated. The lengthscale hyperparameter $\ell$ controls the degree of smoothness, where smaller values allow more rapid variations in the function. The scale parameter $\zeta$ governs the overall magnitude of function variations. More generally, different covariance functions introduce different inductive biases, influencing how the GP extrapolates from observed data.
All covariance functions depend on hyperparameters, collectively denoted as $\bs{\gamma}$, which are typically learned from data by maximizing the marginal log-likelihood (MLL) \cite{williams2006gaussian, wilson2016deep}. The performance of MOBONS relies on well-chosen covariance structures and well-tuned hyperparameters for each function $f_1, \ldots, f_K$. For low- to moderate-dimensional inputs (typically 5–10 dimensions), the RBF or Matern class of covariance functions is often sufficient. For significantly higher-dimensional problems, more expressive covariance functions such as deep kernel learning \cite{wilson2016deep} can be beneficial, as they allow the model to capture complex, non-stationary relationships.

\paragraph{Thompson sampling.} To formulate the acquisition function \eqref{eq:acquisition-func}, we require posterior function samples from the GP models. The complexity of this step depends on the cost of evaluating $\widehat{\bs{F}}_t$. To generate consistent (differentiable) posterior samples, we recommend the efficient procedure in \cite{wilson2020efficiently}, which decomposes a GP posterior sample into two steps: (i) Drawing a realization from the weight-space version of the prior and (ii) Performing a function-space update based on observed data.
This requires knowledge of the weight-space interpretation of the GP, where the covariance function is viewed as an inner product in a feature space. For stationary kernels, this feature space can be approximated using random Fourier features (RFF) \cite{rahimi2007random}, enabling scalable sampling. A detailed discussion of RFF-based posterior sampling is provided in \cite[Section 3]{Bradford2018}.

\paragraph{Approximating the Pareto optimal set.} At each iteration of MOBONS, we solve a multi-objective optimization problem \eqref{eq:acquisition-func} based on GP posterior samples (using the TS paradigm). Although the problem is defined in terms of computationally cheap function samples (unlike the original expensive black-box objectives), it remains a non-convex optimization problem. Consequently, heuristic optimization methods are required for practical implementation.
We adopt NSGA-II \cite{Deb2002}, a widely used evolutionary algorithm that is effective for functions that can be evaluated many times. However, most evolutionary solvers do not natively handle equality or inequality constraints. The most straightforward way to address this is through penalty functions, which adjust the objective values based on constraint violations. A comparison of penalty-based constraint handling techniques is provided in \cite{hobbie2021comparison}.

\paragraph{Selection function.} After computing the (approximate) Pareto optimal set $(X^\star_{t+1}, Y^\star_{t+1})$ in \eqref{eq:acquisition-func}, we must select specific design points for function evaluations. We use the global selection strategy defined in \eqref{eq:global-selection}, though other strategies are also valid.
While \eqref{eq:global-selection} appears computationally challenging, it is efficient in practice due to the finite discrete representation of the Pareto set obtained from NSGA-II (or another solver). Specifically, the selected solution consists of a discrete point set of cardinality $N_\star = | X^\star_{t+1} |$, making the maximin distance calculation straightforward. The required computation involves evaluating $N_\star \times T$ pairwise distances, where $T = | X_t |$ is the number of previously selected points. The final selection involves: (i) Computing all pairwise distances; (ii) Determining the column-wise minimum distance to previously evaluated points; and (iii) Selecting the candidate that maximizes this minimum distance vector. 
This approach ensures good exploration of the design space while maintaining diversity among selected candidates.

\subsection{Handling parallel evaluations and constrained problems}
\label{subsec:parallel-constraints}

In many real-world optimization settings, it is desirable to evaluate multiple candidate solutions in parallel. This situation commonly arises in simulation-based optimization, where multiple computing nodes can evaluate different designs simultaneously, and in experimental settings, where multiple physical tests can be conducted concurrently. The MOBONS framework naturally accommodates parallel evaluations due to the structure of its selection process. Since \eqref{eq:acquisition-func} already produces a set of Pareto-optimal solutions for a given Thompson sample, multiple points can be selected from this set using a greedy maximin strategy based on \eqref{eq:global-selection}. Specifically, after selecting the first candidate $\bs{x}_{t+1}$, the next candidate $\bs{x}_{t+2}$ is chosen as the point that maximizes the worst-case distance to the previously selected points and the first candidate $X_t \cup \{ \bs{x}_{t+1} \}$. This process is repeated until a total of $q$ additional points have been selected. By ensuring that the chosen evaluations are well-distributed, this strategy improves exploration of the objective space while maintaining diversity in the selected points.

If the number of available Pareto-optimal solutions is insufficient to fill all $q$ parallel evaluation slots, additional candidates can be generated by drawing another TS from the function network and re-solving \eqref{eq:acquisition-func}. This approach has been shown to be effective in single-objective Bayesian optimization \cite{kandasamy2018parallelised} and extends naturally to the multi-objective case considered in MOBONS. By leveraging this strategy, the algorithm can efficiently allocate computational resources while maintaining the benefits of its sequential decision-making framework.

Another important consideration in engineering optimization problems is the presence of constraints. Many real-world problems involve inequality constraints that define feasible regions of the design space, such as performance requirements, safety thresholds, or regulatory limits. The function network representation in \eqref{eq:network-functions} used by MOBONS can be readily extended to accommodate constraints by expanding the definition of the network outputs. Specifically, a separate mapping can be introduced from the outputs $\bs{Y}$ to a constraint function $\bs{H}(\bs{x})$, where the constraints are enforced as $\bs{H}(\bs{x}) \leq \bs{0}$. Similar to how the objective functions are obtained through a projection matrix, we define a constraint projection matrix $\bs{C}_\text{cons}$ that maps the outputs to the constraints via the relation $\bs{H}(\bs{x}) = \bs{C}_\text{cons} \bs{Y}^\star(\bs{x})$ where again $\bs{Y}^\star(\bs{x})$ is a solution to \eqref{eq:compact-network} for fixed $\bs{x}$. By appropriately structuring the function network, constraints can always be represented in this form. 

With this formulation, constraints can be incorporated directly into the acquisition function optimization in \eqref{eq:acquisition-func}, ensuring that candidate solutions satisfy feasibility requirements. In practice, constraints are typically handled using penalty functions, as discussed previously, allowing the optimization process to systematically discourage infeasible solutions. Since constraints are treated in a manner analogous to the objectives, MOBONS can efficiently handle constrained problems without requiring fundamental modifications to the underlying framework.

\section{Case Studies}
\label{sec:case-study}

To provide a proof-of-concept evaluation of the performance of MOBONS, we apply it to two case studies: a synthetic test problem based on the well-known ZDT4 benchmark function and a sustainable bioethanol production process. The ZDT4 function represents a challenging multi-objective optimization problem with multiple local minima, making it a useful benchmark for assessing the efficiency of MOBONS in a low-data regime. The second case study involves optimizing the design of an ethanol production process to balance economic performance and environmental impact. In both cases, we compare MOBONS against state-of-the-art multi-objective Bayesian optimization baselines and analyze its ability to efficiently identify high-quality Pareto-optimal solutions.

\subsection{Baseline methods for comparison}

We compare MOBONS against three baseline methods, representing different approaches to multi-objective optimization:
\begin{itemize}
    \item \textbf{Random:} This method selects each query point uniformly at random, serving as a baseline to assess the relative efficiency of more sophisticated methods.
    \item \textbf{Expected Hypervolume Improvement (qEHVI) \cite{Daulton2020}:} This is a fully black-box Bayesian optimization method that selects points to maximize expected improvement in hypervolume, guiding the search towards the Pareto front.
    \item \textbf{Pareto Optimal Thompson Sampling (qPOTS) \cite{Ranganathan2023}:} This is a fully black-box Thompson sampling-based approach for multi-objective optimization. This method can be thought of as a special case of MOBONS that neglects function network structure. In other words, it treats every objective function as an independent black-box function. 
\end{itemize}

To ensure fair comparisons, all baseline methods and MOBONS are implemented using a consistent framework. We use BoTorch \cite{Balandat2020} to implement qEHVI including their default choices for hyperparameters. For qPOTS and MOBONS, we train the GP models using GPyTorch \cite{gardner2018gpytorch} using 5/2 Matern covariance function with automatic relevance determination (ARD) \cite{williams2006gaussian} (meaning individual lengthscale hyperparameters are included for each dimension). We also employ the pymoo \cite{pymoo} implementation of NSGA-II \cite{Deb2002} to numerically optimize the acquisition functions (again using default settings). 

\subsection{Synthetic test problem: ZDT4 benchmark}

The first case study is based on the ZDT4 function \cite{Zitzler2000}, a widely used synthetic benchmark in multi-objective optimization. This function is defined in a ten-dimensional search space ($D=10$), where the first variable defines one objective, while the remaining nine variables introduce nonlinear terms resulting in multiple local minima in the objective function surfaces. These characteristics make ZDT4 particularly challenging for optimization algorithms operating under a limited function evaluation budget.

To introduce function network structure, we model the problem as a system of 11 interconnected nodes, where each function output depends on a subset of inputs. The two competing objectives are defined as follows:
\begin{align}
    y_{1} &= f_{1}({\boldsymbol{x}}) = x_{1}, \\ \notag    
    y_{k} &= f_{k}({\boldsymbol{x}}) = x_{k}^{2} - 10\cos(4\pi x_{k}),~~~ \forall k \in \{2,\dots,10\}, \\ \notag  
    y_{11} &= f_{11}({\boldsymbol{x}})  = \left( 91 + \sum_{k = 2}^{10} y_{k} \right) \left( 1 - \sqrt{\frac{y_{1}}{ 91 + \sum_{k = 2}^{10} y_{k}}} \right), \\ \notag 
    g_{1}({\boldsymbol{x}}) &= y_{1}, \\ \notag 
    g_{2}({\boldsymbol{x}}) &= y_{11}. 
\end{align}
The feasible set for the design variables is constrained within the hyperrectangle: $0 \leq x_1 \leq 1$ and $-10 \leq x_{k} \leq 10$ for $k = 2,\ldots,10$. We assume that the node functions $f_{1,\dots,10}$ are black-box functions (expensive and unknown), while $f_{11}$ is a white-box function (computationally inexpensive and known).

To evaluate algorithm performance, we use hypervolume as the primary metric, measuring the quality of the approximated Pareto front with respect to a reference point $\bs{r}$. Following \cite{Daulton2020}, the hypervolume indicator for an approximate Pareto set $\mathcal{P}$ with a finite number of elements is given by the Lebesgue measure $\lambda_M$ of the space dominated by $\mathcal{P}$ and bounded from below by $\bs{r}$:
\begin{align}
    \text{HV}( \mathcal{P}, \bs{r} ) = \lambda_M \left( \cup_{i=1}^{| \mathcal{P} |} [\bs{r}, \bs{p}_i] \right),
\end{align}
where $[ \bs{r}, \bs{p}_i ]$ is the hyper-rectangle defined by the reference point $\bs{r}$ and a Pareto-optimal solution $\bs{p}_i$. For this case study, we select a reference point of $\bs{r} = [1, 500]$. Each algorithm is initialized with $n = 21$ randomly chosen samples in $\mathcal{X}$, and the total evaluation budget is $N = 121$ (100 additional sequentially selected points). To account for stochastic variability due to random initialization, we repeat all experiments 30 times, ensuring consistent random seeds across all algorithms. 

\begin{figure}[tb]
  \centering
  \includegraphics[width=0.6\textwidth]{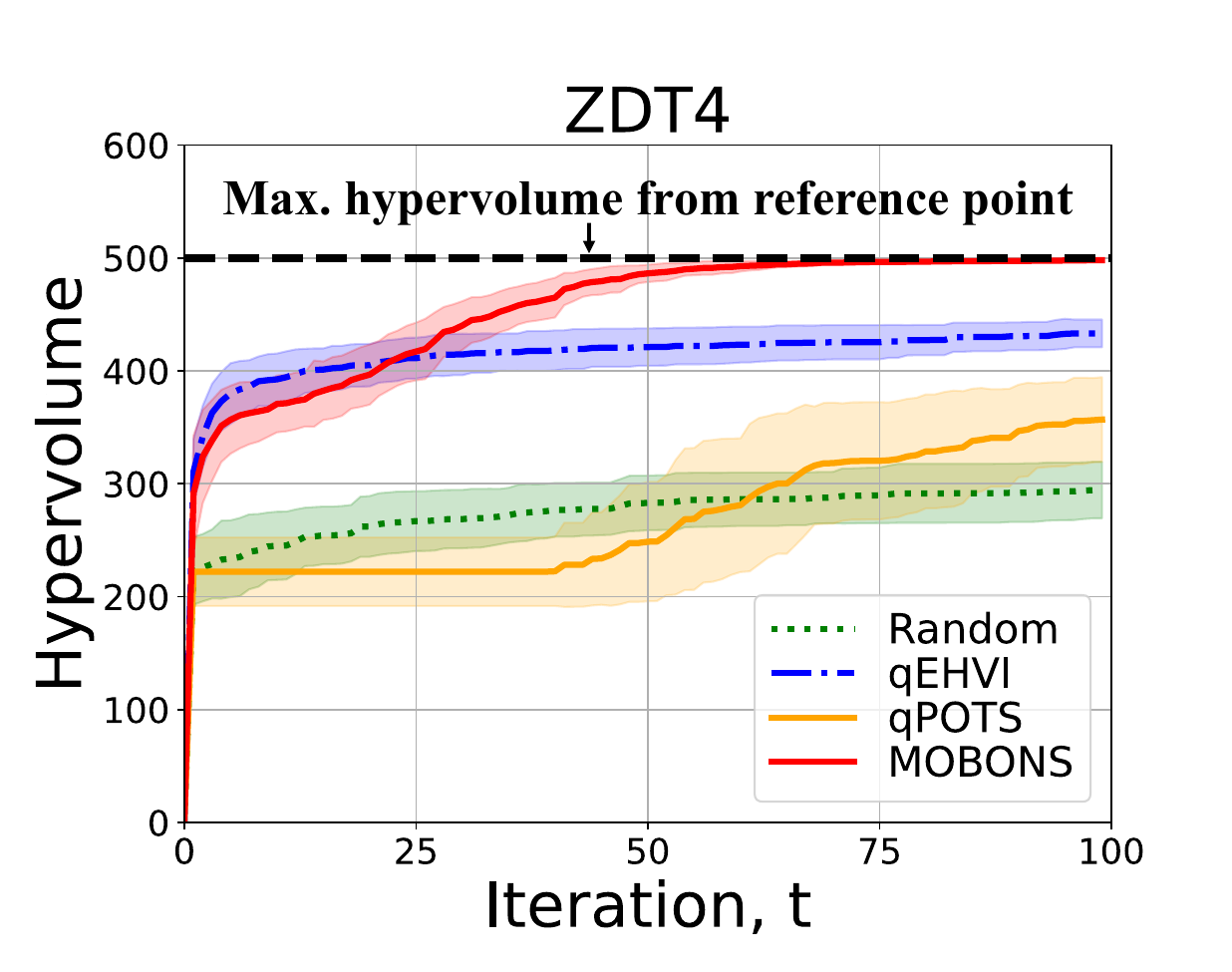}
  \caption{Comparison of multi-objective optimization algorithms on the ZDT4 problem, showing hypervolume progression over iterations. The dashed black line indicates the estimated maximum hypervolume from the reference point. MOBONS reaches the maximum hypervolume faster than all baseline methods, demonstrating its effectiveness in leveraging function network structure when selecting candidates.}
  \label{fig:HV_ZDT4}
\end{figure}

Figure \ref{fig:HV_ZDT4} compares the hypervolume progression of each algorithm. As expected, random sampling exhibits the slowest improvement, providing a baseline for comparison. qPOTS performs moderately well, but it only begins to outperform random sampling after approximately 60 iterations. qEHVI initially demonstrates rapid improvement, quickly increasing the hypervolume, but its progress plateaus after 25 iterations, which is an indication of under-exploration. In contrast, MOBONS maintains steady progress throughout the optimization process, leveraging the function network structure to efficiently navigate the search space. Notably, MOBONS converges to near-optimal hypervolume in approximately 50 iterations, significantly outperforming the baselines. This result highlights the advantage of incorporating network structure in Bayesian optimization, leading to faster convergence and improved Pareto front approximation compared to traditional black-box methods.

\begin{figure}[!htb]
  \centering
  \includegraphics[width=0.8\textwidth]{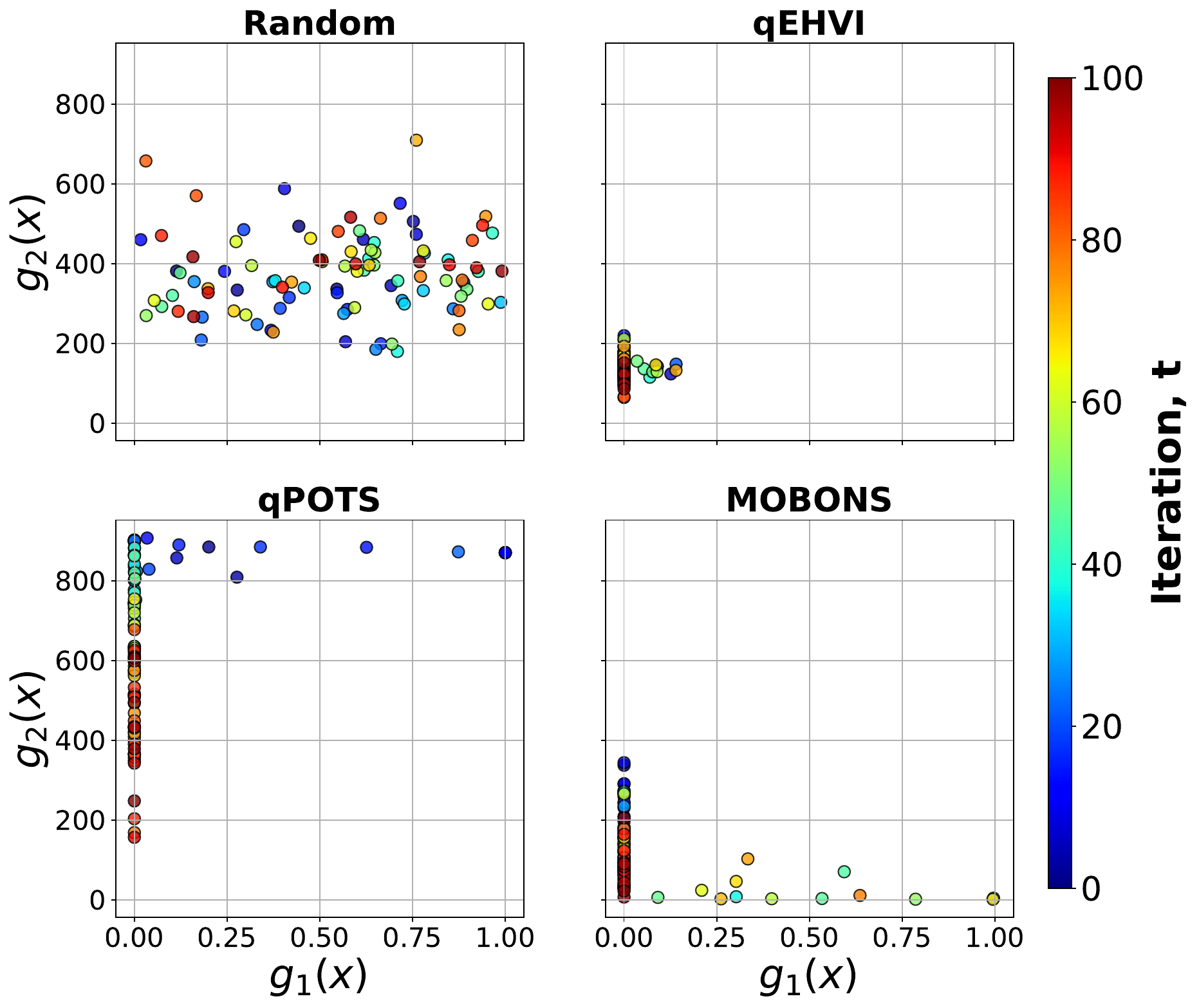}
  \caption{Comparison of function evaluations for each algorithm on the ZDT4 problem, showing the distribution of evaluated points over the two objectives. Each subplot presents the median performing replicate across 30 independent runs. MOBONS successfully pushes samples toward the true Pareto front, demonstrating superior efficiency in balancing exploration and exploitation. qEHVI shows limited exploration, leading to a clustered sampling pattern, while qPOTS and Random fail to fully capture the Pareto-optimal region.}
  \label{fig:ZDT4_MOBONS_scatter}
\end{figure}

To further analyze the distribution of function evaluations across the Pareto front, Figure \ref{fig:ZDT4_MOBONS_scatter} presents scatter plots of the two objective values, $g_1(\bs{x})$ and $g_2(\bs{x})$, obtained over the 100 sequential evaluations for each algorithm in the median-performing replicate. MOBONS clearly achieves a better approximation of the Pareto front, as more samples are pushed toward the lower-left corner of the plot, indicating improved tradeoffs between the objectives. Interestingly, qEHVI, the second-best-performing method in terms of hypervolume, tends to cluster evaluations in certain regions of the objective space, exhibiting limited exploration. In contrast, MOBONS maintains a more diverse distribution of evaluated points, likely contributing to its superior performance in reaching the global Pareto front.

\subsection{Design of sustainable bioethanol process}

The transition to renewable energy sources is a pressing global challenge, driven by the environmental consequences of fossil fuel combustion. Among various alternatives, bioethanol has emerged as a promising renewable fuel due to its production from biomass feedstocks via fermentation. Unlike fossil fuels, ethanol combustion can be partially carbon-neutral, as CO$_2$ emissions are offset by the CO$_2$ uptake of the biomass feedstocks during growth. However, large-scale ethanol production still presents sustainability challenges, including energy-intensive separation processes and inefficient feedstock utilization. To address these issues, optimization methods must balance economic and environmental trade-offs to develop cost-effective and sustainable ethanol production systems.

\begin{figure}[tb]
\centering
\includegraphics[width=0.7\textwidth]{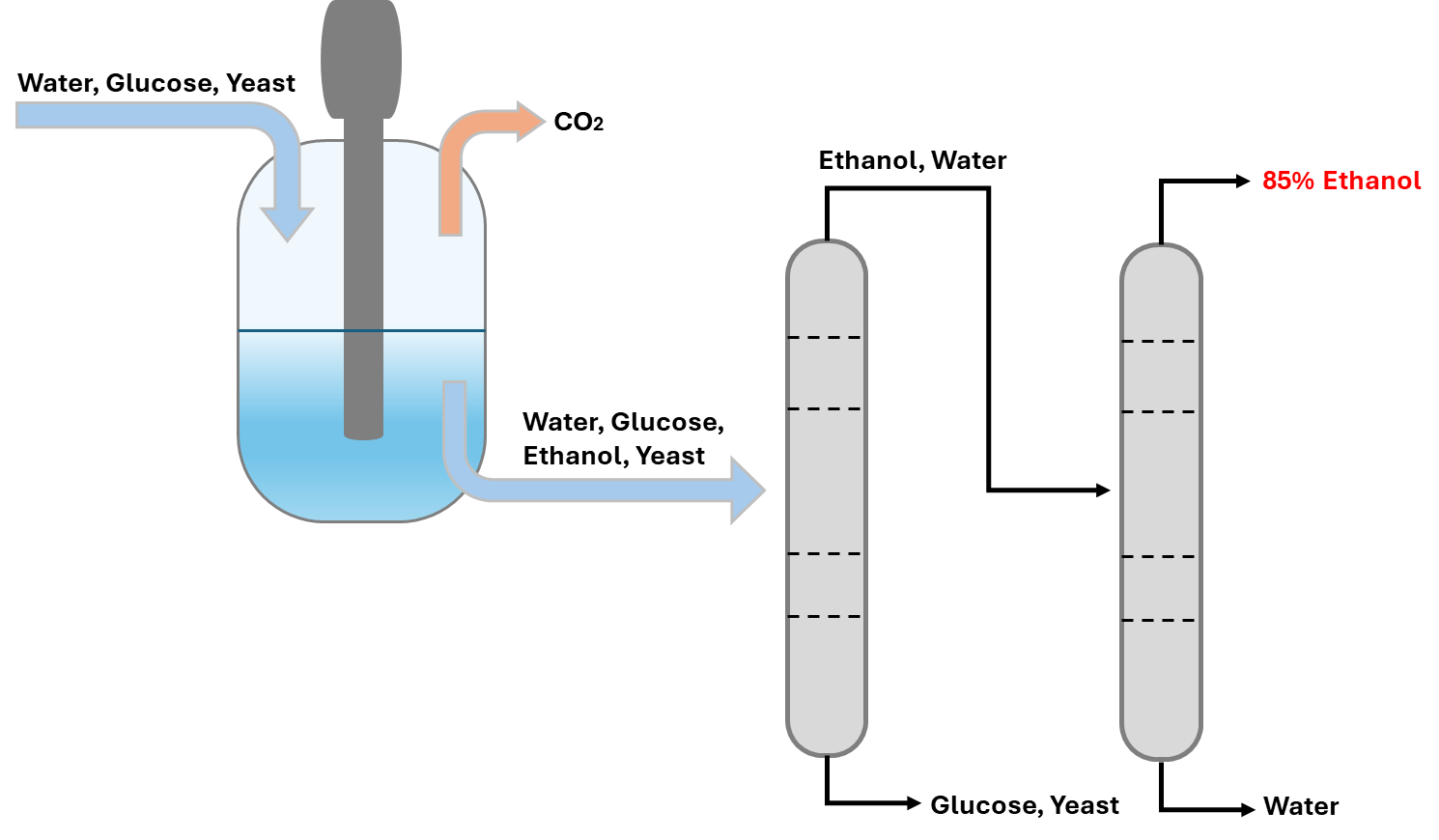}
\caption{Process diagram for the ethanol fermentation system, consisting of a fermentation reactor followed by two distillation columns. The optimization goal is to balance economic and environmental performance by adjusting process design variables.}
\label{fig:process_diagram}
\end{figure}

\paragraph{Process description and implementation.} In this case study, we apply MOBONS to optimize a steady-state bioethanol production process with competing economic and environmental objectives. This example serves as a proof of concept, demonstrating the potential of MOBONS in process systems engineering. The process consists of a fermentation reactor followed by two distillation columns, as shown in Figure~\ref{fig:process_diagram}. The simulation is implemented using BioSTEAM \cite{CortesPea2020}, an open-source biochemical process simulation platform that enables the evaluation of complex thermodynamic and economic trade-offs.

\paragraph{Problem formulation and function network representation.} We consider eight decision variables, including the feed flowrates of water and glucose ($F$), the reaction temperature and pressure of the fermentation reactor ($T_{\text{rxn}}, P_{\text{rxn}}$), the operating pressures of the two distillation columns ($P_1, P_2$), the reflux-to-minimum-reflux ratios for both distillation columns ($RR_1, RR_2$), and the ethanol purity in the distillate of the first distillation column ($p$). The bounds on these variables are listed below at the top of Table~\ref{tab:decision var Pareto}. 
The two competing objectives are:
\begin{itemize}
\item Economic performance, measured by revenue generated from high-purity ethanol sales. This metric is maximized. 
\item Environmental impact, quantified using global warming potential (GWP), which accounts for CO$_2$ emissions. This metric is minimized. 
\end{itemize}
It is important to note that MOBONS can be applied to any user-defined objective functions; the choices above serve as reasonable illustrative examples.

\begin{figure}[tb]
\centering
\includegraphics[width=0.95\textwidth]{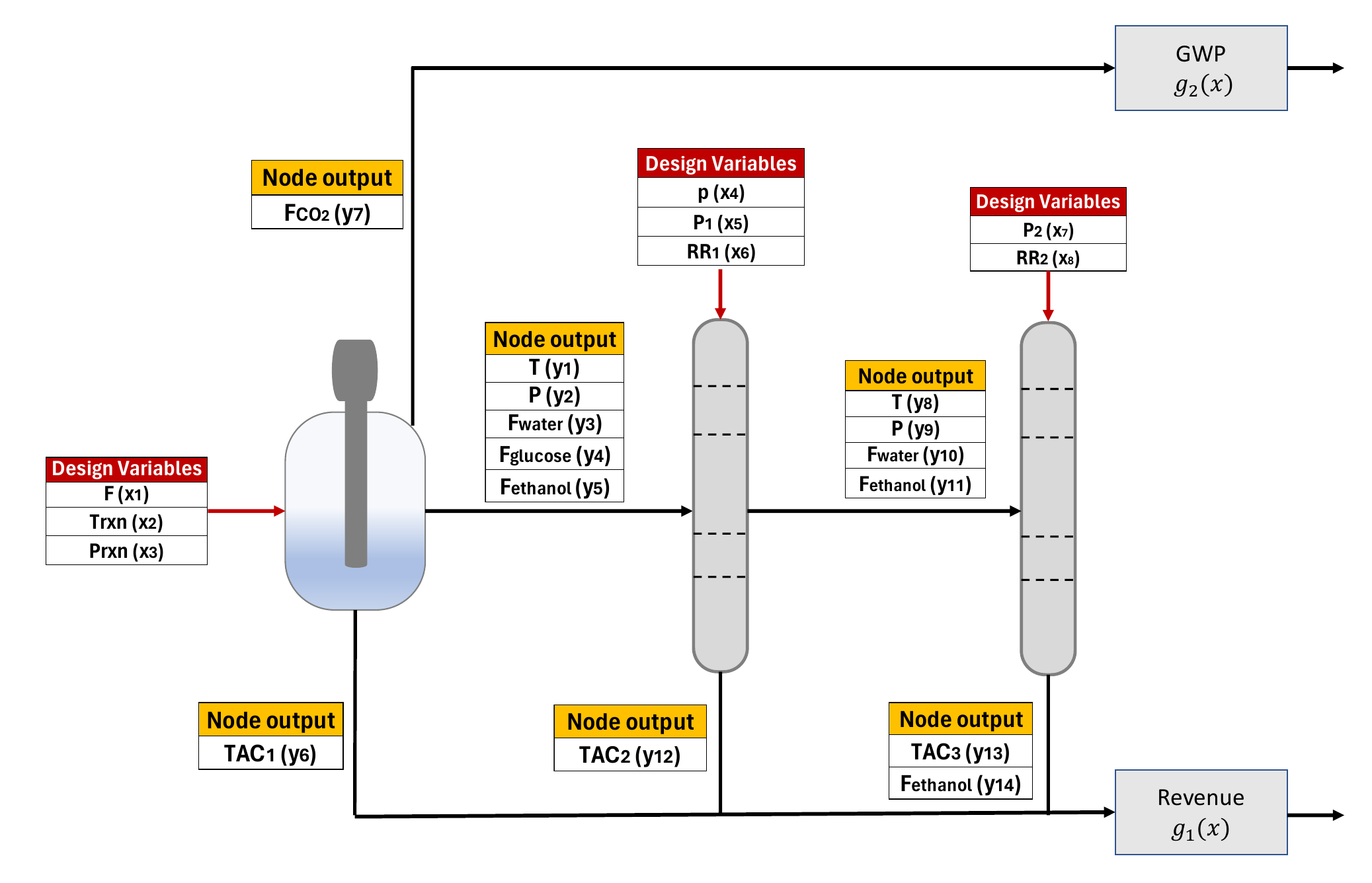}
\caption{Function network representation of the ethanol fermentation process, showing decision variables (red), intermediate outputs (orange), and objective functions (gray). Revenue and GWP are modeled as explicit white-box calculations, leveraging known process relationships to improve optimization efficiency.}
\label{fig:process function network}
\end{figure}

MOBONS models this ethanol production system as a function network, depicted in Figure~\ref{fig:process function network}. The network consists of (i) Decision variables (red boxes) that define the process configuration; (ii) Intermediate process outputs (orange boxes) that include reaction and separation unit responses; and (iii) Objective functions (gray boxes), where the final calculations of revenue and GWP are explicitly modeled as white-box calculations. Thus, the network consists of a total of $K=5$ node, with 3 being black-box and 2 being white-box function nodes. To facilitate reproducibility, we provide an open-source implementation of this case study on Github at the following link: \href{https://github.com/PaulsonLab/MOBONS}{https://github.com/PaulsonLab/MOBONS}. Detailed descriptions of the white-box equations can be found in the provided code.

\begin{figure}[tb]
\centering
\includegraphics[width=0.7\textwidth]{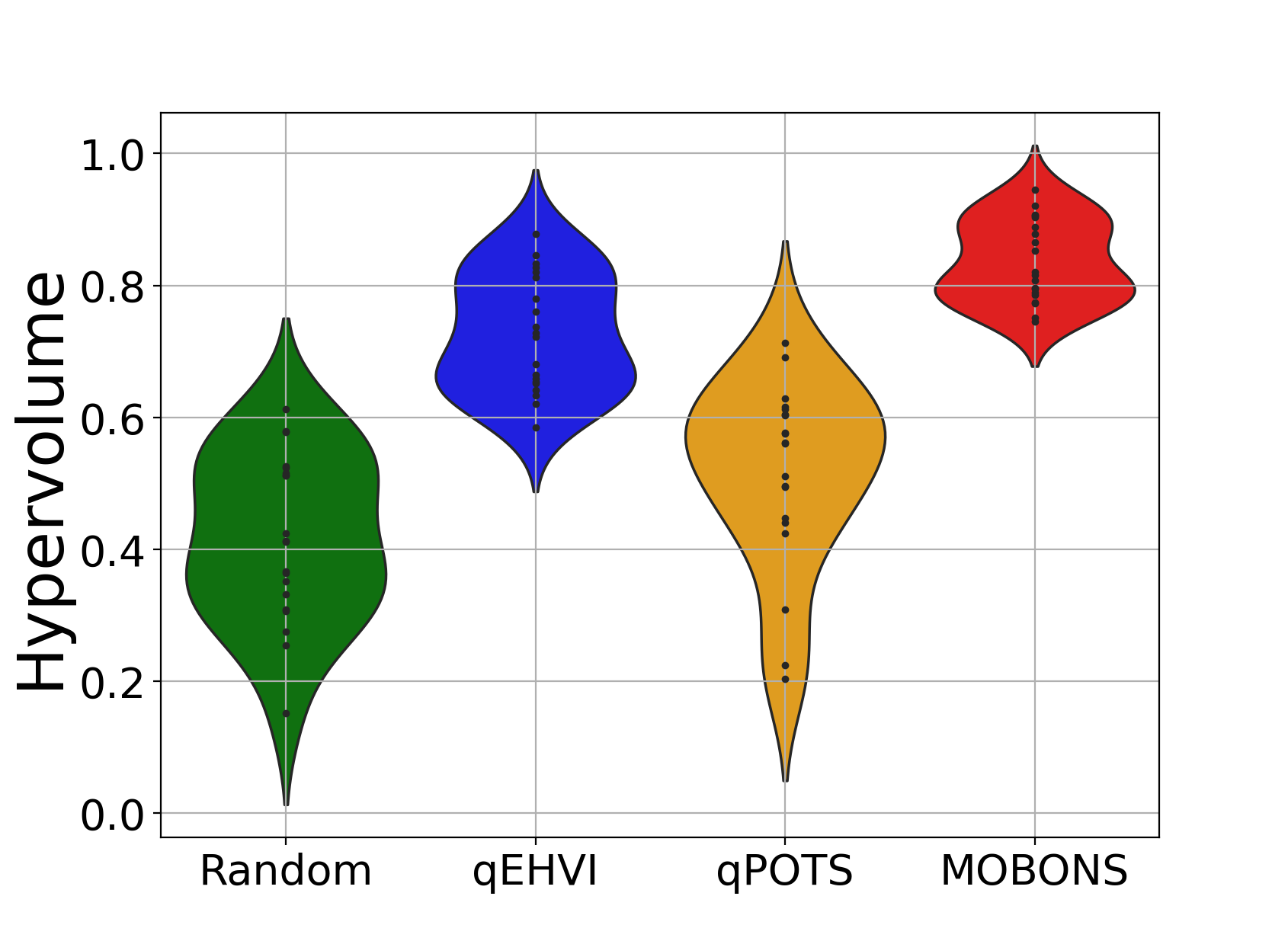}
\caption{Comparison of hypervolume achieved by different multi-objective optimization methods across 20 independent runs. MOBONS consistently outperforms qPOTS and qEHVI, demonstrating its ability to efficiently explore and exploit the search space.}
\label{fig:violin}
\end{figure}

\begin{figure}[tb]
\centering
\includegraphics[width=0.7\textwidth]{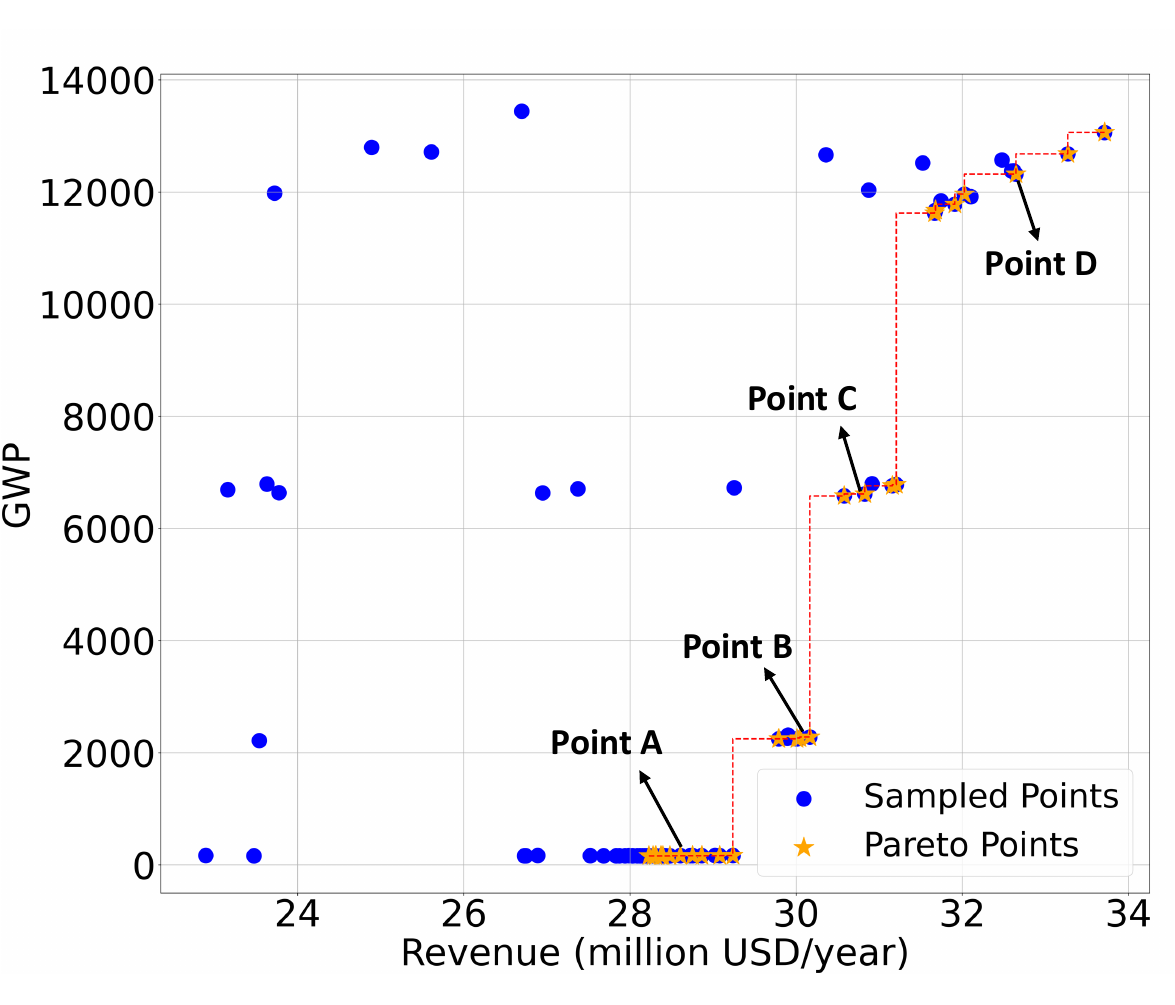}
\caption{Pareto front and design points identified by MOBONS. The front exhibits four distinct regions, highlighting different trade-offs between revenue and environmental impact. The discontinuities arise due to the stepwise penalties applied to GWP as CO$_2$ emissions increase.}
\label{fig:Pareto scatter}
\end{figure}

\paragraph{Optimization performance and hypervolume analysis.} As in the previous case study, we compare MOBONS against three baseline algorithms. Each algorithm is initialized with 20 random samples and allowed a total budget of 70 function evaluations. The optimization results are evaluated using the hypervolume metric, which quantifies the volume dominated by the obtained Pareto front relative to a predefined reference point $\bs{r} = [28.3, 13600]$ (where the first element is revenue in units of million USD, and the second element is GWP). 
Figure~\ref{fig:violin} presents the hypervolume distributions across 20 optimization replicates for each method. MOBONS consistently achieves the highest hypervolume, demonstrating its ability to efficiently identify high-quality Pareto-optimal solutions. Figure~\ref{fig:Pareto scatter} shows the Pareto front identified by MOBONS in one of the best optimization runs. Notably, the Pareto front exhibits four distinct clusters, reflecting different trade-off regimes between revenue and GWP.

\begin{table}[tb]
    \centering
    \caption{Bounds and optimal values at the four Pareto optimal solutions shown in Figure \ref{fig:Pareto scatter} for all decision variables in the bioethanol process design problem.}
    \begin{tabular}{|c|c|c|c|c|c|c|c|c|} \hline 
         &  $F$ (kmol/hr)&  $T_{\text{rxn}}$ ($^{\circ}\text{C}$)&  $P_{\text{rxn}}$ (atm)&  $P_1$ (atm)&  $RR_1$&  $P_2$ (atm)&  $RR_2$& $p$\\ \hline 
         Range &  $[90,110]$ &  $[30,40]$ &  $[1,5]$ & $[1,5]$  &  $[0.1,10]$ &$[1,5]$  &  $[2,10]$ & $[0.5,0.7]$ \\ \hline
         Point A&  90.5&  30.0&  5.0&  4.0&  6.9&  3.7&  2.0& 0.7\\ \hline 
         Point B&  95.4&  30.6&  5.0&  4.6&  5.4&  2.5&  2.0& 0.7\\ \hline 
         Point C&  98.2&  35.1&  4.9&  5.0&  1.8&  2.5&  3.0& 0.7\\ \hline
 Point D& 101.0& 36.7& 5.0& 2.3& 5.1& 3.3& 3.5&0.7\\\hline
    \end{tabular}
    \label{tab:decision var Pareto}
\end{table}

\begin{figure}[tb]
\centering
\includegraphics[width=1.0\textwidth]{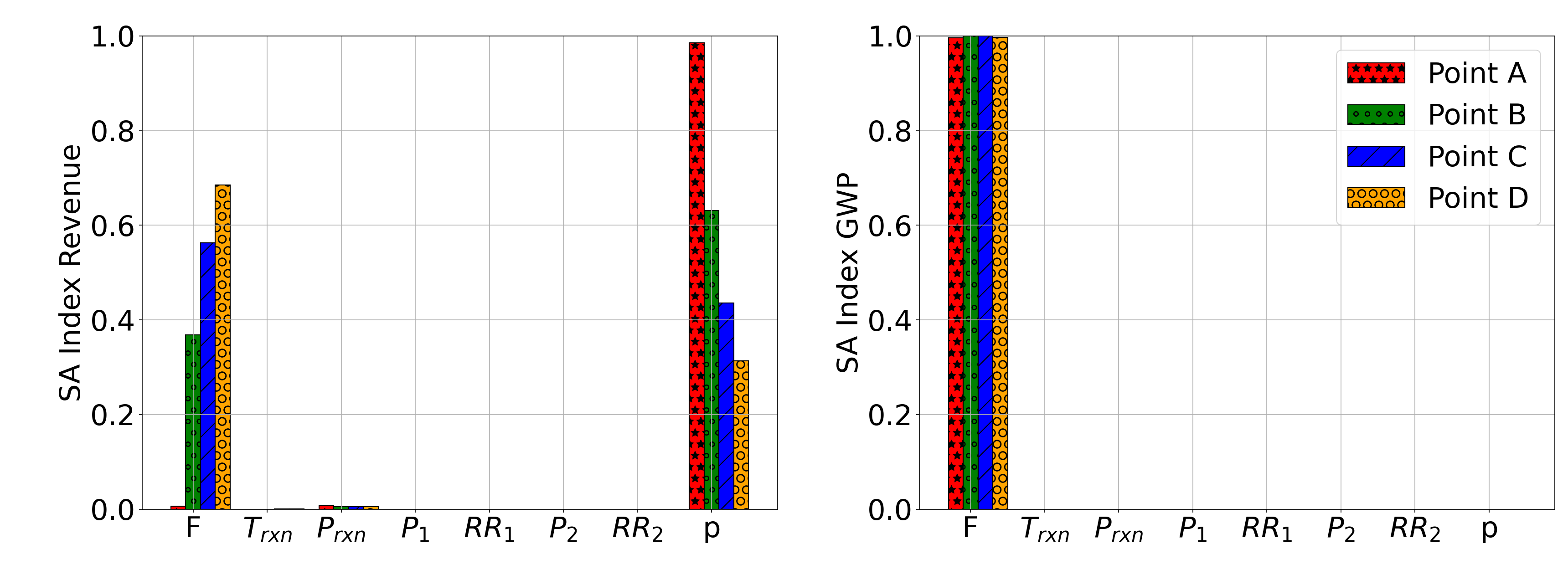}
\caption{Sobol sensitivity analysis results for revenue (left) and GWP (right) across four Pareto-optimal points shown in Figure \ref{fig:Pareto scatter}. Revenue is primarily influenced by water/glucose feedflow rate ($F$) and ethanol purity in the first distillation column ($p$), while GWP is only sensitive to $F$.}
\label{fig:sensitivity analysis}
\end{figure}

\paragraph{Local sensitivity analysis.} Another advantage of MOBONS is that it produces a surrogate model that incorporates network structure, enabling additional analyses such as sensitivity analysis (SA). Here, we use SA to examine the impact of different decision variables on the two objectives (revenue and GWP). Since the function network GP model may not be globally accurate, we focus on a local SA around solutions along the Pareto frontier. We conduct local SA on four selected Pareto solutions (points A, B, C, and D in Figure~\ref{fig:Pareto scatter}) by perturbing each decision variable within a $\pm 5$\% range. To improve model accuracy in these local regions, we generate an additional 24 function evaluations centered around each Pareto point. First-order Sobol indices \cite{sobol2001global} are then computed to quantify the relative influence of each decision variable on revenue and GWP. Figure~\ref{fig:sensitivity analysis} presents the results of the local SA. We see that revenue is primarily influenced by water/glucose feed flowrate ($F$) and ethanol purity in the first distillation column ($p$). Notably, as revenue increases along the Pareto front (moving from left to right), its sensitivity to $F$ grows, while its sensitivity to $p$ diminishes. On the other hand, we see that GWP is predominantly impacted by $F$, with negligible (local) sensitivity to other variables. This result aligns with expectations, as CO$_2$ emissions scale directly with the amount of material fed into the process. 

One of the significant advantages of this type of approach is that it enables SA in a way that would be virtually impossible with a purely black-box optimization framework. In traditional black-box optimization, obtaining the Pareto front is already computationally expensive, and conducting local SA at key points on the front would require an impractically large number of additional function evaluations. By leveraging the function network GP model learned by MOBONS, we can efficiently approximate the local response of the system to small perturbations in decision variables, allowing us to gain meaningful insights at a fraction of the computational cost. It is worth emphasizing that this analysis focuses on local sensitivity around selected Pareto-optimal solutions rather than a global sensitivity analysis across the entire design space. While global SA can provide a broader understanding of variable importance, it is often infeasible for high-dimensional, computationally expensive problems. In contrast, local SA, as performed here, is still valuable for decision-making because it highlights which variables have the most influence on system performance near optimal designs. This helps designers (or stakeholders) more quickly identify bottlenecks, assess trade-offs, and estimate design variable importance. 

\section{Conclusion}
\label{sec:conclusions}

In this chapter, we presented a flexible and general framework, \textbf{M}ulti-\textbf{O}bjective \textbf{B}ayesian \textbf{O}ptimization for \textbf{N}etwork \textbf{S}ystems (MOBONS), for handling complex, networked models involving expensive black-box function evaluations. Unlike most existing approaches that rely either on fully equation-oriented (white-box) representations or treat the entire problem as a monolithic black-box, our method leverages a network (graph) structure to capture key interactions and information flows between subsystem functions. While MOBONS builds on prior work that uses function networks in Bayesian optimization, it provides two primary advances: to our knowledge, it is the first algorithm to address both multi-objective problems and function networks simultaneously, and it is also the first to accommodate cyclic graphs (e.g., recycle streams). These features are incorporated in a manner that keeps the algorithm both computationally efficient and straightforward to implement.

We provided detailed implementation guidelines and practical considerations, including strategies for incorporating constraints on network outputs and for supporting batch (parallel) evaluations when multiple resources are available. In addition, we supply supporting code to facilitate the application of MOBONS in diverse engineering and scientific problems. Our numerical results, encompassing both a synthetic test system and a green ethanol production case study, illustrate the value of exploiting structural information in networked models. Even in relatively high-dimensional settings, MOBONS identifies Pareto-optimal solutions that effectively balance competing objectives, such as economic performance and environmental impact. We also demonstrate how the resulting function network surrogate model can be used to perform sensitivity analysis around the identified Pareto frontier designs, offering valuable insights for decision makers.

Although these results are promising, the present work serves primarily as a proof of concept, leaving several avenues open for future research. First, a more rigorous theoretical analysis of MOBONS’s convergence properties -- particularly in the presence of cycles and complex constraints -- would help establish broader guarantees. Second, the efficiency of the multi-objective subproblem could be further enhanced by developing specialized solvers capable of handling both equality and inequality constraints more effectively. Third, exploring more sophisticated covariance modeling approaches (such as deep kernel learning or node-specific kernels) may capture each node function’s unique characteristics more precisely. Finally, the impact of different selection rules for choosing candidate points from the approximate Pareto front warrants systematic study, given the well-known exploration-exploitation tradeoff. We hope these directions will inspire future developments and encourage the broader adoption of network-aware Bayesian optimization methods in both industry and academia.

\section*{Acknowledgments}

The authors gratefully acknowledge the support from the Ohio Department of Higher Education Harmful Algal Bloom Research Initiative (ODHE HABRI). 

\bibliographystyle{unsrturl}
\bibliography{references}

\begin{thebibliography}{10}

\bibitem{boulding1966economics}
Kenneth~E. Boulding.
\newblock The economics of the coming spaceship earth, 1966.

\bibitem{Bakshi_nature_book}
Bhavik~R. Bakshi, editor.
\newblock {\em Engineering and Ecosystems: Seeking Synergies Toward a
  Nature-Positive World}.
\newblock Springer International Publishing, 2023.
\newblock URL: \url{http://dx.doi.org/10.1007/978-3-031-35692-6}, \href
  {https://doi.org/10.1007/978-3-031-35692-6}
  {\path{doi:10.1007/978-3-031-35692-6}}.

\bibitem{EPA_website}
{U.S. Environmental Protection Agency}.
\newblock U.s. environmental protection agency - scientific and technical
  information, 2025.
\newblock URL:
  \url{https://cfpub.epa.gov/si/si_public_record_report.cfm?Lab=NERL&dirEntryId=319430}.

\bibitem{Takamatsu1970}
Takeichiro Takamatsu, Iori Hashimoto, and Hiromu Ohno.
\newblock Optimal design of a large complex system from the viewpoint of
  sensitivity analysis.
\newblock {\em Industrial \& Engineering Chemistry Process Design and
  Development}, 9(3):368–379, July 1970.
\newblock URL: \url{http://dx.doi.org/10.1021/i260035a004}, \href
  {https://doi.org/10.1021/i260035a004} {\path{doi:10.1021/i260035a004}}.

\bibitem{Nishida1974}
N.~Nishida, A.~Ichikawa, and E.~Tazaki.
\newblock Synthesis of optimal process systems with uncertainty.
\newblock {\em Industrial \& Engineering Chemistry Process Design and
  Development}, 13(3):209–214, July 1974.
\newblock URL: \url{http://dx.doi.org/10.1021/i260051a003}, \href
  {https://doi.org/10.1021/i260051a003} {\path{doi:10.1021/i260051a003}}.

\bibitem{Grossmann1978}
I.~E. Grossmann and R.~W.~H. Sargent.
\newblock Optimum design of chemical plants with uncertain parameters.
\newblock {\em AIChE Journal}, 24(6):1021–1028, November 1978.
\newblock URL: \url{http://dx.doi.org/10.1002/aic.690240612}, \href
  {https://doi.org/10.1002/aic.690240612} {\path{doi:10.1002/aic.690240612}}.

\bibitem{Halemane1983}
K.~P. Halemane and I.~E. Grossmann.
\newblock Optimal process design under uncertainty.
\newblock {\em AIChE Journal}, 29(3):425–433, May 1983.
\newblock URL: \url{http://dx.doi.org/10.1002/aic.690290312}, \href
  {https://doi.org/10.1002/aic.690290312} {\path{doi:10.1002/aic.690290312}}.

\bibitem{Swaney1985}
R.~E. Swaney and I.~E. Grossmann.
\newblock An index for operational flexibility in chemical process design. part
  i: Formulation and theory.
\newblock {\em AIChE Journal}, 31(4):621–630, April 1985.
\newblock URL: \url{http://dx.doi.org/10.1002/aic.690310412}, \href
  {https://doi.org/10.1002/aic.690310412} {\path{doi:10.1002/aic.690310412}}.

\bibitem{Grossmann1987}
I.E. Grossmann and C.A. Floudas.
\newblock Active constraint strategy for flexibility analysis in chemical
  processes.
\newblock {\em Computers \& Chemical Engineering}, 11(6):675–693, January
  1987.
\newblock URL: \url{http://dx.doi.org/10.1016/0098-1354(87)87011-4}, \href
  {https://doi.org/10.1016/0098-1354(87)87011-4}
  {\path{doi:10.1016/0098-1354(87)87011-4}}.

\bibitem{Narraway1991}
L.T. Narraway, J.D. Perkins, and G.W. Barton.
\newblock Interaction between process design and process control: economic
  analysis of process dynamics.
\newblock {\em Journal of Process Control}, 1(5):243–250, November 1991.
\newblock URL: \url{http://dx.doi.org/10.1016/0959-1524(91)85015-B}, \href
  {https://doi.org/10.1016/0959-1524(91)85015-b}
  {\path{doi:10.1016/0959-1524(91)85015-b}}.

\bibitem{Luyben1994}
M.L. Luyben and C.A. Floudas.
\newblock Analyzing the interaction of design and control—1. a multiobjective
  framework and application to binary distillation synthesis.
\newblock {\em Computers \& Chemical Engineering}, 18(10):933–969, October
  1994.
\newblock URL: \url{http://dx.doi.org/10.1016/0098-1354(94)E0013-D}, \href
  {https://doi.org/10.1016/0098-1354(94)e0013-d}
  {\path{doi:10.1016/0098-1354(94)e0013-d}}.

\bibitem{Burnak2019}
Baris Burnak, Nikolaos~A. Diangelakis, and Efstratios~N. Pistikopoulos.
\newblock Towards the grand unification of process design, scheduling, and
  control—utopia or reality?
\newblock {\em Processes}, 7(7):461, July 2019.
\newblock URL: \url{http://dx.doi.org/10.3390/pr7070461}, \href
  {https://doi.org/10.3390/pr7070461} {\path{doi:10.3390/pr7070461}}.

\bibitem{Niederer2021}
Steven~A. Niederer, Michael~S. Sacks, Mark Girolami, and Karen Willcox.
\newblock Scaling digital twins from the artisanal to the industrial.
\newblock {\em Nature Computational Science}, 1(5):313–320, May 2021.
\newblock URL: \url{http://dx.doi.org/10.1038/s43588-021-00072-5}, \href
  {https://doi.org/10.1038/s43588-021-00072-5}
  {\path{doi:10.1038/s43588-021-00072-5}}.

\bibitem{Sharma2022}
Angira Sharma, Edward Kosasih, Jie Zhang, Alexandra Brintrup, and Anisoara
  Calinescu.
\newblock Digital twins: State of the art theory and practice, challenges, and
  open research questions.
\newblock {\em Journal of Industrial Information Integration}, 30:100383,
  November 2022.
\newblock URL: \url{http://dx.doi.org/10.1016/j.jii.2022.100383}, \href
  {https://doi.org/10.1016/j.jii.2022.100383}
  {\path{doi:10.1016/j.jii.2022.100383}}.

\bibitem{Kushner1964}
H.~J. Kushner.
\newblock A new method of locating the maximum point of an arbitrary multipeak
  curve in the presence of noise.
\newblock {\em Journal of Basic Engineering}, 86(1):97–106, March 1964.
\newblock URL: \url{http://dx.doi.org/10.1115/1.3653121}, \href
  {https://doi.org/10.1115/1.3653121} {\path{doi:10.1115/1.3653121}}.

\bibitem{Mokus1975}
J.~Močkus.
\newblock {\em On {Bayesian} methods for seeking the extremum}, page 400–404.
\newblock Springer Berlin Heidelberg, 1975.
\newblock URL: \url{http://dx.doi.org/10.1007/3-540-07165-2_55}, \href
  {https://doi.org/10.1007/3-540-07165-2_55}
  {\path{doi:10.1007/3-540-07165-2_55}}.

\bibitem{Hickman2022}
Riley~J. Hickman, Matteo Aldeghi, Florian H\"{a}se, and Alán Aspuru-Guzik.
\newblock Bayesian optimization with known experimental and design constraints
  for chemistry applications.
\newblock {\em Digital Discovery}, 1(5):732–744, 2022.
\newblock URL: \url{http://dx.doi.org/10.1039/D2DD00028H}, \href
  {https://doi.org/10.1039/d2dd00028h} {\path{doi:10.1039/d2dd00028h}}.

\bibitem{Kudva2022}
Akshay Kudva, Farshud Sorourifar, and Joel~A. Paulson.
\newblock Constrained robust {Bayesian} optimization of expensive noisy
  black‐box functions with guaranteed regret bounds.
\newblock {\em AIChE Journal}, 68(12), August 2022.
\newblock URL: \url{http://dx.doi.org/10.1002/aic.17857}, \href
  {https://doi.org/10.1002/aic.17857} {\path{doi:10.1002/aic.17857}}.

\bibitem{banerjee2010computationally}
Ipsita Banerjee, Siladitya Pal, and Spandan Maiti.
\newblock Computationally efficient black-box modeling for feasibility
  analysis.
\newblock {\em Computers \& Chemical Engineering}, 34(9):1515--1521, 2010.

\bibitem{rogers2015feasibility}
Amanda Rogers and Marianthi Ierapetritou.
\newblock Feasibility and flexibility analysis of black-box processes part 2:
  Surrogate-based flexibility analysis.
\newblock {\em Chemical Engineering Science}, 137:1005--1013, 2015.

\bibitem{Geremia2023}
Margherita Geremia, Fabrizio Bezzo, and Marianthi~G. Ierapetritou.
\newblock A novel framework for the identification of complex feasible space.
\newblock {\em Computers \& Chemical Engineering}, 179:108427, November 2023.
\newblock URL: \url{http://dx.doi.org/10.1016/j.compchemeng.2023.108427}, \href
  {https://doi.org/10.1016/j.compchemeng.2023.108427}
  {\path{doi:10.1016/j.compchemeng.2023.108427}}.

\bibitem{Kudva2024}
Akshay Kudva, Wei-Ting Tang, and Joel~A. Paulson.
\newblock Robust {Bayesian} optimization for flexibility analysis of expensive
  simulation-based models with rigorous uncertainty bounds.
\newblock {\em Computers \& Chemical Engineering}, 181:108515, February 2024.
\newblock URL: \url{http://dx.doi.org/10.1016/j.compchemeng.2023.108515}, \href
  {https://doi.org/10.1016/j.compchemeng.2023.108515}
  {\path{doi:10.1016/j.compchemeng.2023.108515}}.

\bibitem{Paulson2024BO4Sustaibalilty}
Joel~A Paulson and Calvin Tsay.
\newblock Bayesian optimization as a flexible and efficient design framework
  for sustainable process systems.
\newblock {\em Current Opinion in Green and Sustainable Chemistry}, page
  100983, 2024.

\bibitem{Boukouvala2016}
Fani Boukouvala and Christodoulos~A. Floudas.
\newblock Argonaut: Algorithms for global optimization of constrained grey-box
  computational problems.
\newblock {\em Optimization Letters}, 11(5):895–913, April 2016.
\newblock URL: \url{http://dx.doi.org/10.1007/s11590-016-1028-2}, \href
  {https://doi.org/10.1007/s11590-016-1028-2}
  {\path{doi:10.1007/s11590-016-1028-2}}.

\bibitem{Beykal2018}
Burcu Beykal, Fani Boukouvala, Christodoulos~A. Floudas, and Efstratios~N.
  Pistikopoulos.
\newblock Optimal design of energy systems using constrained grey-box
  multi-objective optimization.
\newblock {\em Computers \& Chemical Engineering}, 116:488–502, August 2018.
\newblock URL: \url{http://dx.doi.org/10.1016/j.compchemeng.2018.02.017}, \href
  {https://doi.org/10.1016/j.compchemeng.2018.02.017}
  {\path{doi:10.1016/j.compchemeng.2018.02.017}}.

\bibitem{paulson2022cobalt}
Joel~A Paulson and Congwen Lu.
\newblock {COBALT: COnstrained Bayesian optimizAtion of computationaLly
  expensive grey-box models exploiting derivaTive information}.
\newblock {\em Computers \& Chemical Engineering}, 160:107700, 2022.

\bibitem{Kudva2022draco}
Akshay Kudva, Farshud Sorouifar, and Joel~A. Paulson.
\newblock Efficient robust global optimization for simulation-based problems
  using decomposed {G}aussian processes: Application to {MPC} calibration.
\newblock In {\em 2022 American Control Conference (ACC)}, pages 2091--2097,
  2022.
\newblock \href {https://doi.org/10.23919/ACC53348.2022.9867777}
  {\path{doi:10.23919/ACC53348.2022.9867777}}.

\bibitem{Schweidtmann2021}
Artur~M. Schweidtmann, Dominik Bongartz, Daniel Grothe, Tim Kerkenhoff,
  Xiaopeng Lin, Jaromił Najman, and Alexander Mitsos.
\newblock Deterministic global optimization with gaussian processes embedded.
\newblock {\em Mathematical Programming Computation}, 13(3):553–581, June
  2021.
\newblock URL: \url{http://dx.doi.org/10.1007/s12532-021-00204-y}, \href
  {https://doi.org/10.1007/s12532-021-00204-y}
  {\path{doi:10.1007/s12532-021-00204-y}}.

\bibitem{Lu2023}
Congwen Lu and Joel~A. Paulson.
\newblock No-regret constrained {Bayesian} optimization of noisy and expensive
  hybrid models using differentiable quantile function approximations.
\newblock {\em Journal of Process Control}, 131:103085, November 2023.
\newblock URL: \url{http://dx.doi.org/10.1016/j.jprocont.2023.103085}, \href
  {https://doi.org/10.1016/j.jprocont.2023.103085}
  {\path{doi:10.1016/j.jprocont.2023.103085}}.

\bibitem{lu2025bo4io}
Yen-An Lu, Wei-Shou Hu, Joel~A Paulson, and Qi~Zhang.
\newblock Bo4io: A {Bayesian} optimization approach to inverse optimization
  with uncertainty quantification.
\newblock {\em Computers \& Chemical Engineering}, 192:108859, 2025.

\bibitem{openfoam}
OpenFOAM Foundation.
\newblock Openfoam: The open source computational fluid dynamics {(CFD)}
  toolbox, 2025.
\newblock Accessed: 2025-02-03.
\newblock URL: \url{https://www.openfoam.org}.

\bibitem{Guinee2002}
Jeroen~B. Guinée.
\newblock {\em Handbook on Life Cycle Assessment: Operational Guide to the ISO
  Standards}.
\newblock Springer, Dordrecht, 2002.
\newblock \href {https://doi.org/10.1007/BF02978784}
  {\path{doi:10.1007/BF02978784}}.

\bibitem{iTree}
i~Tree.
\newblock i-tree eco (version 6.0.32) [computer software], 2023.
\newblock Accessed: 2025-02-03.
\newblock URL: \url{https://www.itreetools.org/i-tree-tools-download}.

\bibitem{DWSIM}
{Daniel Medeiros}.
\newblock Dwsim.
\newblock URL: \url{http://dwsim.inforside.com.br}.

\bibitem{astudillo2019bayesian}
Raul Astudillo and Peter Frazier.
\newblock Bayesian optimization of composite functions.
\newblock In {\em International Conference on Machine Learning}, pages
  354--363. PMLR, 2019.

\bibitem{BOFN}
Raul Astudillo and Peter Frazier.
\newblock Bayesian optimization of function networks.
\newblock {\em Advances in Neural Information Processing Systems},
  34:14463--14475, 2021.

\bibitem{BOFN_PE}
Poompol Buathong, Jiayue Wan, Raul Astudillo, Sam Daulton, Maximilian Balandat,
  and Peter~I. Frazier.
\newblock {B}ayesian optimization of function networks with partial
  evaluations.
\newblock In Ruslan Salakhutdinov, Zico Kolter, Katherine Heller, Adrian
  Weller, Nuria Oliver, Jonathan Scarlett, and Felix Berkenkamp, editors, {\em
  Proceedings of the 41st International Conference on Machine Learning}, volume
  235 of {\em Proceedings of Machine Learning Research}, pages 4752--4784.
  PMLR, 21--27 Jul 2024.
\newblock URL: \url{https://proceedings.mlr.press/v235/buathong24a.html}.

\bibitem{gonzalez2024bois}
Leonardo~D Gonz{\'a}lez and Victor~M Zavala.
\newblock {BOIS: Bayesian optimization of interconnected systems}.
\newblock {\em IFAC-PapersOnLine}, 58(14):446--451, 2024.

\bibitem{gonzalez2025implementation}
Leonardo~D Gonz{\'a}lez and Victor~M Zavala.
\newblock Implementation of a {Bayesian} optimization framework for
  interconnected systems.
\newblock {\em Industrial \& Engineering Chemistry Research}, 2025.

\bibitem{khoshraftar2024survey}
Shima Khoshraftar and Aijun An.
\newblock A survey on graph representation learning methods.
\newblock {\em ACM Transactions on Intelligent Systems and Technology},
  15(1):1--55, 2024.

\bibitem{kandasamy2018parallelised}
Kirthevasan Kandasamy, Akshay Krishnamurthy, Jeff Schneider, and Barnab{\'a}s
  P{\'o}czos.
\newblock Parallelised {B}ayesian optimisation via thompson sampling.
\newblock In {\em International Conference on Artificial Intelligence and
  Statistics}, pages 133--142. PMLR, 2018.

\bibitem{kudva2025bonsai}
Akshay Kudva and Joel~A Paulson.
\newblock {BONSAI: Structure-exploiting robust Bayesian optimization for
  networked black-box systems under uncertainty}.
\newblock {\em Computers \& Chemical Engineering}, page 109393, 2025.

\bibitem{mavrotas2009effective}
George Mavrotas.
\newblock Effective implementation of the $\varepsilon$-constraint method in
  multi-objective mathematical programming problems.
\newblock {\em Applied Mathematics and Computation}, 213(2):455--465, 2009.

\bibitem{hart2011pyomo}
William~E Hart, Jean-Paul Watson, and David~L Woodruff.
\newblock Pyomo: modeling and solving mathematical programs in python.
\newblock {\em Mathematical Programming Computation}, 3(3):219--260, 2011.

\bibitem{Deb2002}
K.~Deb, A.~Pratap, S.~Agarwal, and T.~Meyarivan.
\newblock A fast and elitist multiobjective genetic algorithm: Nsga-ii.
\newblock {\em IEEE Transactions on Evolutionary Computation}, 6(2):182--197,
  2002.
\newblock \href {https://doi.org/10.1109/4235.996017}
  {\path{doi:10.1109/4235.996017}}.

\bibitem{Daulton2020}
Samuel Daulton, Maximilian Balandat, and Eytan Bakshy.
\newblock Differentiable expected hypervolume improvement for parallel
  multi-objective {Bayesian} optimization.
\newblock 2020.
\newblock URL: \url{https://arxiv.org/abs/2006.05078}, \href
  {https://doi.org/10.48550/ARXIV.2006.05078}
  {\path{doi:10.48550/ARXIV.2006.05078}}.

\bibitem{Bradford2018}
Eric Bradford, Artur~M. Schweidtmann, and Alexei Lapkin.
\newblock Efficient multiobjective optimization employing gaussian processes,
  spectral sampling and a genetic algorithm.
\newblock {\em Journal of Global Optimization}, 71(2):407–438, February 2018.
\newblock URL: \url{http://dx.doi.org/10.1007/s10898-018-0609-2}, \href
  {https://doi.org/10.1007/s10898-018-0609-2}
  {\path{doi:10.1007/s10898-018-0609-2}}.

\bibitem{Ranganathan2023}
S.~Ashwin Renganathan and Kade~E. Carlson.
\newblock qpots: Efficient batch multiobjective {Bayesian} optimization via
  pareto optimal {Thompson} sampling, 2023.
\newblock URL: \url{https://arxiv.org/abs/2310.15788}, \href
  {https://doi.org/10.48550/ARXIV.2310.15788}
  {\path{doi:10.48550/ARXIV.2310.15788}}.

\bibitem{williams2006gaussian}
Christopher~KI Williams and Carl~Edward Rasmussen.
\newblock {\em Gaussian processes for machine learning}, volume~2.
\newblock MIT press Cambridge, MA, 2006.

\bibitem{sun2019synthesizing}
Furong Sun, Robert~B Gramacy, Benjamin Haaland, Siyuan Lu, and Youngdeok Hwang.
\newblock Synthesizing simulation and field data of solar irradiance.
\newblock {\em Statistical Analysis and Data Mining: The ASA Data Science
  Journal}, 12(4):311--324, 2019.

\bibitem{wilson2016deep}
Andrew~Gordon Wilson, Zhiting Hu, Ruslan Salakhutdinov, and Eric~P Xing.
\newblock Deep kernel learning.
\newblock In {\em Artificial Intelligence and Statistics}, pages 370--378.
  PMLR, 2016.

\bibitem{wilson2020efficiently}
James Wilson, Viacheslav Borovitskiy, Alexander Terenin, Peter Mostowsky, and
  Marc Deisenroth.
\newblock Efficiently sampling functions from {G}aussian process posteriors.
\newblock In {\em International Conference on Machine Learning}, pages
  10292--10302. PMLR, 2020.

\bibitem{rahimi2007random}
Ali Rahimi and Benjamin Recht.
\newblock Random features for large-scale kernel machines.
\newblock {\em Advances in neural information processing systems}, 20, 2007.

\bibitem{hobbie2021comparison}
Jared~G Hobbie, Amir~H Gandomi, and Iman Rahimi.
\newblock A comparison of constraint handling techniques on {NSGA-II}.
\newblock {\em Archives of Computational Methods in Engineering},
  28(5):3475--3490, 2021.

\bibitem{Balandat2020}
Maximilian Balandat, Brian Karrer, Daniel~R. Jiang, Samuel Daulton, Benjamin
  Letham, Andrew~Gordon Wilson, and Eytan Bakshy.
\newblock Botorch: A framework for efficient monte-carlo {Bayesian}
  optimization.
\newblock 2019.
\newblock URL: \url{https://arxiv.org/abs/1910.06403}, \href
  {https://doi.org/10.48550/ARXIV.1910.06403}
  {\path{doi:10.48550/ARXIV.1910.06403}}.

\bibitem{gardner2018gpytorch}
Jacob~R Gardner, Geoff Pleiss, David Bindel, Kilian~Q Weinberger, and
  Andrew~Gordon Wilson.
\newblock Gpytorch: Blackbox matrix-matrix gaussian process inference with gpu
  acceleration.
\newblock In {\em Advances in Neural Information Processing Systems}, 2018.

\bibitem{pymoo}
J.~{Blank} and K.~{Deb}.
\newblock pymoo: Multi-objective optimization in python.
\newblock {\em IEEE Access}, 8:89497--89509, 2020.

\bibitem{Zitzler2000}
Eckart Zitzler, Kalyanmoy Deb, and Lothar Thiele.
\newblock Comparison of multiobjective evolutionary algorithms: Empirical
  results.
\newblock {\em Evolutionary Computation}, 8(2):173–195, June 2000.
\newblock URL: \url{http://dx.doi.org/10.1162/106365600568202}, \href
  {https://doi.org/10.1162/106365600568202}
  {\path{doi:10.1162/106365600568202}}.

\bibitem{CortesPea2020}
Yoel Cortes-Peña, Deepak Kumar, Vijay Singh, and Jeremy~S. Guest.
\newblock Biosteam: A fast and flexible platform for the design, simulation,
  and techno-economic analysis of biorefineries under uncertainty.
\newblock {\em ACS Sustainable Chemistry \& Engineering}, 8(8):3302–3310,
  January 2020.
\newblock URL: \url{http://dx.doi.org/10.1021/acssuschemeng.9b07040}, \href
  {https://doi.org/10.1021/acssuschemeng.9b07040}
  {\path{doi:10.1021/acssuschemeng.9b07040}}.

\bibitem{sobol2001global}
Ilya~M Sobol.
\newblock Global sensitivity indices for nonlinear mathematical models and
  their monte carlo estimates.
\newblock {\em Mathematics and computers in simulation}, 55(1-3):271--280,
  2001.

\end{thebibliography}

\end{document}